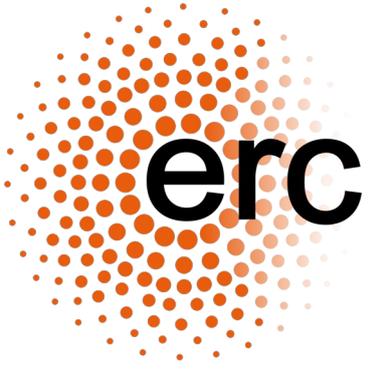
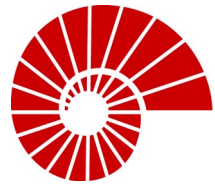

# Global Contentious Politics Database (GLOCON) Annotation Manuals

Fırat Duruşan, Ali Hürriyetoğlu, Erdem Yörük, Osman Mutlu, Çağrı Yoltar, Burak Gürel, Alvaro Comin

Version 1.01

2022-05-11

# Global Contentious Politics Database (GLOCON) Annotation Manuals

Developed By

Fırat Duruşan, Ali Hürriyetoğlu, Erdem Yörük, Osman Mutlu, Çağrı Yoltar, Burak Gürel, Alvaro Comin

# Acknowledgments

This work is funded by the European Research Council (ERC) Starting Grant awarded to Dr. Erdem Yörük for the project Emerging Markets Welfare (project ID 714868). The research project is hosted by the Koç University and has benefited from the rich intellectual environment created by its distinguished academic staff throughout its five year span. The linguistic annotation process also benefited from research collaboration between Koç University and the University of São Paulo, each lending excellent teams of young social scientists who have provided high quality annotations in English, Portuguese and Spanish languages, as well as their apt suggestions and insights in preparing the manual. For such invaluable contributions, the authors of this manual are grateful to Gizem Türkaslan, Ezgi Özçelik, Bağlan Deniz, Balacan Fatıma Ayar, Eylem Taylan, Pelin Kılınçarslan, Sercan Taş, Enes Sarı, Ricardo Framil Filho, Mariana dos Santos, Rodrigo Brandão, Patricia Jimenez, Táli Pires de Almeida, Tamires Cristina da Silva, Juliana Machado, Ivan Vieira, Sümercan Bozkurt, Ömer Özcan, and Selim Kırılmaz.

# Table of Contents

# 1 Introduction

Emerging Markets Welfare project[1] investigates the effects of contentious politics on welfare state programs in countries of the Global South. It hypothesizes that government response to social contention is a significant factor that shapes welfare policies. It is in this respect that mapping the dynamics of social contention in a given country becomes crucial, and duly constitutes a fundamental component of the entire project. Investigating the causal relationship between social contention and government policy involves more than a simple correlation, particularly if the focus is on specific government action, namely welfare policies. The map of social contention adequate for such an understanding should thus go beyond laying out basic trends of ebbing and flowing of social contention over space and time and provide insight into particularities such as the types of action repertoires, levels of violence, characteristics of actors or social groups that engage in contentious politics, the characteristics of the demands that they raise.

The purpose of the second work package of the EMW Project is to draw the aforementioned map of social contention. For achieving this purpose, we created a database of contentious politics events through the extraction of information from the news reports that are featured in the most prominent online sources each focus country has to offer. The Global Contentious Politics Database (GLOCON)[2] records contentious politics events (referred to as protest events for the sake of brevity) that take place within the borders of our focus countries with all the information available in the source about the events' time and place, actor, type, demands raised, violence level. As of the moment, the GLOCON database contains protest event data from India, China, South Africa, Argentina, and Brazil. It features data in three languages: English for India, China, and South Africa data, Spanish for Argentina data, and Portuguese for Brazil data. The database was created in a way that is able to accommodate additions of other focus countries and/or news sources in the future.

The database creation utilized automated text processing tools that detects if a news article contains a protest event, locate protest information within the article, and extract pieces of information regarding the detected protest events. The basis of training and testing the automated tools is the GLOCON Gold Standard Corpus (GSC), which contains news articles from multiple sources from each focus country. The articles in the GSC were manually coded by skilled annotators in both classification and extraction tasks with the utmost accuracy and consistency that automated tool development demands. In order to assure these, the annotation manuals[3] in this document lay out the rules according to which annotators code the news articles. Annotators refer to the manuals at all times for all annotation tasks and apply the rules that they contain.

---

[1] Yo can find more information on the project at https://emw.ku.edu.tr/. You can stay updated by following the twitter account at https://twitter.com/EmergingWelfare and the youtube channel at https://www.youtube.com/channel/UC1SDR9yjXAFTAGRVHSyefuw
[2] The database can be found at https://glocon.ku.edu.tr/.
[3] This manual was published along with the publication Hürriyetoğlu, A., Yörük, E., Mutlu, O., Duruşan, F., Yoltar, Ç., Yüret, D. and Gürel, B.: Cross-context news corpus for protest event-related knowledge base construction. Data Intelligence 3(2), 2021. doi: 10.1162/dint_a_00092. See the final section for the list of publications from the EMW Project that refer to it. Regarding the contents of the manual, please contact Fırat Duruşan (email: fdurusan@ku.edu.tr).

Despite the EMW Project's focus on the countries of the Global South, and the initial choice of a limited number of countries to be featured in the GSC, none of the rules or principles contained in this manual is more or less applicable to certain countries, sources or periods than others. The GLOCON database aims to be inclusive and capable of expanding. Securing consistency, reliability, and validity of data in the face of temporal and spatial expansion requires that annotation principles are generally applicable and that they are applied consistently.

The annotation process is composed of three main levels for each news report document. The document-level annotation determines the news articles that contain information on actual (past or ongoing) protest events. The sentence-level annotation aims to locate sentences that contain protest event-related information. In the final phase, words or phrases that give concrete information about protest events are detected.

Corresponding to the document and sentence classification, and information extraction tasks, there are three main and two supplemental manuals which together cover the entire annotation process from the document, through the sentence, to the token level. The first manual is the Document Level Protest Annotation Manual (DOLPAM) which establishes the rules for determining news articles that contain protest events; in other words, classifying news articles into those which contain protest events and those which do not. It lays out the protest event ontology, that is, the protest event definition which specifies the range of contentious politics events that are included in the scope of the project. It introduces and exemplifies different types of protest events, and defines the criteria to which a news report must conform to be labeled as a protest event article. The following Sentence Level Protest Annotation Manual (SELPAM) carries on with classifying the sentences in the documents that have already been classified as protest event articles. Similarly, it defines and exemplifies event sentences and enumerates the rules by which sentences are labeled as event sentences and non-event sentences. The third and final main manual is the Token Level Protest Annotation Manual (TOLPAM) which is the longest and most detailed of the three main level manuals. It defines the types of event-related information that the project aims at collecting from news articles and explains how expressions within the event sentences which contain these pieces of information are tagged. The remaining two manuals are supplemental manuals that label further information about the events that are already extracted in the three main levels of annotation. Both define annotation tasks that are performed on the document level. The first is the Violent Protest Events Annotation Manual which lays out the rules for classifying news reports that contain protest events into categories of violent and non-violent. The following, Protest Event Demands Annotation Manual aims at setting the rules for labeling the demands and/or grievances associated with the protest events that are extracted in the news articles. More detailed information about each manual can be found under their respective headings.

Even though every particular level of annotation has its respective annotation manual, the whole process must be thought of as an integrated whole as each level of annotation is premised on the results of the previous level. Hence, familiarizing oneself with all the manuals before starting annotation on any single level is recommended. Knowing in advance what the sentence and token level annotation tasks entail would help an annotator working on the document level considerably. That said, it is neither practical nor advisable to try to learn all annotation procedures by heart. Memory is prone to mislead, and recurrent reference to the manuals is the preferred way of utilizing them. Thus, annotators must read the entire manual before starting annotation, and remember to refer to it when there is the slightest doubt about a rule or a difficult case.

The content of the annotation manual is built on the general principles and standards of linguistic annotation laid out in other prominent annotation manuals such as ACE, CAMEO, and TimeML. These principles, however, have been adapted or rather modified heavily to accommodate the social scientific concepts and variables employed in the EMW project. The manual has been molded throughout a long trial and error process that accompanied the annotation of the GSC. It owes much of its current shape to the meticulous work and invaluable feedback provided by highly specialized teams of annotators, whose diligence and expertise greatly increased the quality of the corpus.

# 2 Document Level Protest Annotation Manual

The protest event classification task on the document level aims at determining which news articles contain contentious politics events. This manual will define the event ontology, that is the meaning and range of protest events, and how it should be decided that a news article contains such an event. All the news articles that are processed in this task will be classified into "protest" and "no protest" categories in the annotation interface we utilize for this task. Please remember that annotation decisions must be based exclusively on the text that is being processed and the rules laid out in annotation manuals. Although this point might sound obvious, there will be times when the annotator will have prior knowledge about the events mentioned in news articles before reading them, and it is only natural that such knowledge tends to play a role in the annotator's decision. This will all the more be the case as annotation progresses and the annotator keeps gaining knowledge about the case countries. The focus must always remain on what the specific news article that is being annotated contains as information rather than prior or background knowledge of the annotator. Please read carefully and comprehend the event ontology below, and remember to come back to this manual whenever you encounter news articles that prove hard to decide.

## 2.1 Identifying Protest Events

The event we simply refer to as protest events are those events that are comprised within the scope of contentious politics. Contentious politics events are politically motivated collective action events that lay outside the official mechanisms of political participation associated with formal government institutions of the country in which the said action takes place. The term contentious politics is so broad in its coverage of very diverse collective action types that it might seem counter-intuitive to group them under the same heading. In order to lay out this diversity, which the simple definition of contentious politics given above hardly reveals, different collective action, and event types will be identified below. Contentious politics can be said to comprise three broad categories of collective actions and four categories of events. Remember that it is not necessary on this level to distinguish the categories from each other and pigeonhole the events encountered in news articles into one of them. This enumeration is done simply to establish the limits of the protest event ontology we operationalize.

### 2.1.1 Types of collective action

#### 2.1.1.1 Political mobilizations

Political mobilizations refer to instances of collective action where political parties, organizations, or individual representatives of these entities mobilize any group of participants for political goals, demands, or grievances. Such political organizers might be in government or opposition in their respective political contexts.

### 2.1.1.2 Social Protests

The second class of collective action that contentious politics encompasses social protests in the narrow sense. A protest action is one through which individuals, groups, or organizations voice their objections, oppositions, demands, or grievances to a person or institution of authority.

### 2.1.1.3 Group Confrontations

The final broad category of collective action we consider within the scope of contentious politics is incidents of social confrontation where two or more social groups engage in political action against each other. Various sources of disagreement might exist between social groups, such as conflicting economic interests, ethnic, religious, or gender-based identity differences, or ideological disagreements. Any instance of an actual confrontation between groups based on such grounds is within the scope of our understanding of contentious politics. Such confrontations might or might not involve state and/or political institutions.

## 2.1.2 Types of Events

These broad categories of contentious politics manifest themselves in different types of events. Again, with the aim of simplification, four broad event type categories can be identified: demonstrations, industrial actions, group clashes, and politically motivated violent actions targeting officials (including security personnel) or civilians. Note that there might be certain types of protest that are unique to a particular country, or a protest might have a unique name in a particular country. Please refer to the supervisors or country experts when you encounter a type of protest which is not covered in this manual.

### 2.1.2.1 Demonstrations

A demonstration is defined as a form of political action in which a demand and/or grievance is raised outside the given institutionalized forms of political participation in a country. The event aims to draw the attention of politicians and/or the general public to said goal by making it visible in the public sphere which implies any space which is open to members of the public. Violent or peaceful forms of social protest that take place in open or closed public spaces such as streets, plazas, vicinity of prominent buildings fall in the category of demonstrations. Notable examples of demonstrations are as follows: Protest marches, silent marches, rallies, demonstrations, outdoor press declarations, gatherings, sit-ins, acts of civil disobedience, dharnas, bandhs, demonstrators clashing with security forces, commemorations and religious rituals (when they become means of protest, e.g. akhand paths), collective petitions (collecting signatures), shouting slogans, carrying banners, self-immolation (and threats of self-immolation), hunger strikes, fasts, barricading, picketing (roadblocks), burning of vehicles, occupations of public buildings that are not included in the industrial action category and the like.

### 2.1.2.2 Industrial Actions

Industrial actions are types of protest events that take place within workplaces and/or involve the production process in the protest. Notable examples include any kind of strike (comprising slowdown strikes, wildcat strikes, sympathy strikes, and green bans), workplace occupations, boycotts, picket lines, and gheraoes.

### 2.1.2.3 Group clashes

Group clash events are instances of confrontation that stem from politicized conflicts (e.g. identity or economic interest-based or ideological conflicts) between social groups such as fighting, lynching, ransacking, arson, any armed or unarmed clash between civilians of a political nature. Actions that target religious, ethnic or similar minorities fall within this category. Said actions can be unidirectional, that is, the target of the event might or might not retaliate during the event.

### 2.1.2.4 Political violence and militancy

Finally, politically motivated violent events fall within our event definition. These actions are usually carried out by political and/or militant organizations which resort to violence, and target officials or civilians. Notable examples of this type of event are kidnapping, assassination, bombing, suicide bombing, hijacking, and the like. Even though such actions are generally carried out by organizations, there might be cases in which such acts are carried out by individuals not necessarily affiliated with organizations or groups. Hence, if the act which intentionally resorts to violence carries political goals, they are included in our definition even when they are carried out by a proverbial "lone wolf".

## 2.1.3 Necessary characteristics of events to be annotated

Having identified the types of contentious politics and associated event types, laying out necessary criteria in terms of their reporting in the news articles is in order. The protest event mentions that are encountered in the articles must have the characteristics enumerated below so that the article can be classified as a protest article.

### 2.1.3.1 The Necessity of civilian actors

Firstly, the participants or organizers of protest events must include at least one non-state actor. Non-state actors might be political or non-political organizations such as parties, associations, or trade unions, or they might simply be individuals or a group of citizens such as residents of a locality or employees of a certain workplace.

### 2.1.3.2 Time and location necessities

The most important characteristics of an event are its time and place. We expect events to have concrete times and places as these are among the most important pieces of information we aim to collect when we analyze the selected event-related articles in the information extraction phase. Below you will find an abstract definition of these necessities. To better come to grips with these principles, you can refer to section 1.5.2. below, where they are elaborated in more concrete terms. Time and location are pieces of information that make it certain that the event-mention in the news article is that of a specific event that actually took place rather than a conceptual or general reference.

Most essentially, the event to be labeled must certainly have taken place. This means that only the past or ongoing events are to be included. The most certain indicator of this is the tense the article uses in narrating the story. The events that are in scope are most commonly narrated in the past tense. For ongoing events, the narrator might prefer to use the simple present or continuous tenses which also indicate the certainty of the event's occurrence.

Secondly, we expect events to be current affairs and event articles to provide a piece of relatively specific time information that leaves no doubt as to the event's occurrence. Definite time indicators such as complete dates, days, hours, or time expressions such as yesterday, last week, etc. would certainly qualify as time indicators. We also allow vague time indicators which inform us that the event is a current affair as in "Last month's Naxal violence in the region" or "recent riots in Gujarat".

We expect concrete information about the location of the event to be present in the article. The location information might be the place of an event, that is, a geographical location more specific than the country where the event took place such as state, city, town, district, village, etc. If such a place is not explicitly mentioned, the facility information, that is, information regarding the type of space where the event takes place (such as a building, street, plaza, stadium, etc.) will also be sufficient. However, some protest actions such as those that take place in cyberspace do not have physical locations. Likewise, certain other events do not take place in specific locations, such as product boycotts and signature campaigns. These and similar events which, by their nature, do not take place in specific physical locations must be annotated as protest events despite lacking location information.

## 2.1.4 Examples of cases that should be considered protest events

In this section, several specific protest event cases that are in our scope will be enumerated and examples will be provided. Note that the cases below are far from exhaustive in terms of our event definition. They will merely provide several instances where the inclusion of the event in the article is not immediately obvious. You might think of them as corner cases of sorts.

  i. Sometimes protest events exist in the news articles even though they are not mentioned in the title or are not the main topic of the article. Put differently, we are not primarily concerned with the title or main topic of the articles that we analyze, but rather whether they contain a protest event or not.

  ii. Political parties or their leaders mobilizing groups of people are manifestly included in our scope. This might look less obvious when said politicians or parties are in the government and therefore the whole affair might look official or less of a civilian character. We include these and all cases where the event is actively participated by members of the public. An example would be the political party mobilizing its supporters for an election rally or during an official ceremony celebrating a national day. Similarly, political party activities which aim at mass participation such as mass feedings or youth festivals are venues for mass mobilizations and thus are included in our event definition.

iii. Protest actions by members of political parties, including members of parliament, are included if these actions take place publicly, and outside the state institutions and normal procedures of government. The actions that take place without public participation within the confines of state institutions, such as legislative bodies, that are unusual and/or not associated with the normal operation of government are included as protest events. Events like hunger strikes or armed violence that are manifestly unusual in terms of the normal functioning of state institutions are accepted as protest events even when they physically take place within state institutions.

iv. Events that are part of armed conflicts between states and (non-state) organizations are included regardless of the scale of the conflict, provided that they are initiated by non-state actors. Clashes between state and non-state actors which are initiated by state actors are included if it is actively responded to by the non-state actor. Military coups are exceptions to this rule (see the following section).

v. Though rare, there are instances of state officials acting outside their official capacity and protesting. Public employees striking or engaging in work-related protests are usual and easy to recognize, but some subtle cases might involve them appearing to carry out their duties. Be on the lookout for acts of insubordination or misconduct that are carried out as advancing an explicitly stated political agenda that belongs to those who act. Cases in which security personnel attack civilians or refuse to prevent acts of violence among civilians might be difficult to identify as protest events as they might be associated with official policies of state repression.

vi. Sometimes it might not be immediately obvious that an act of violence, such as a case of murder or a clash between groups has a political character which would make the case a protest event in our understanding. Assassinations of or attacks against prominent public figures, be it political figures or professionals of public stature (high-level bureaucrats, lawyers, writers, etc.), are very likely to be protest events. They should be treated as protest events unless it is certain that they happen to be caused by personal or non-political reasons.

vii. Group clashes based on economic interests might not have an obvious political angle and look like instances of petty turf wars. Exemplary cases include fights between Uber and taxi drivers or clashes between different fishermen communities. These and similar clashes are very much included in our scope.

viii. We expect the included events to have a certainty of occurrence as mentioned above. But in certain instances, the threat or attempt of certain actions have the effect of a protest in and of themselves despite not being carried out for one reason or another. Threats of

- violent actions such as assassination, bombing, attack, and self-immolation are such cases that have significant effects that make them eligible for inclusion in our event definition, even if they are not carried out or somehow obstructed from being fulfilled.

ix. Non-political contexts can well become scenes of political protests. Sports events, concerts, and even religious rituals might become scenes or media of protest actions. Keep in mind that the protest events that take place in such non-political settings are also in our scope, provided that they have political motivations or target authorities.

x. The events that are alleged by civilian or official actors to have taken place in the news reports are included even when the said claims are not verified. E.g. "The police have reportedly rounded up 11 members of the BKU for allegedly attacking one Kishan Singh"

xi. Certain protest events are not space-bound, that is to say, they do not take place in a specific physical space by definition. These are regarded as an exception to the location necessity rule. Protest actions that take place on the world wide web that are sometimes referred to under the name of "hacktivism" such as instances of cyber-attacks of a political nature are among these and will be labeled as events. Also, events like product boycotts, sometimes referred to as consumer activism, do not take place in specific locations and must be labeled as protest events despite not meeting the location requirement.

xii. Collective petitions and declarations that are open to members of the public for participation as signatories are included as events.

xiii. Sometimes, riot events might be reported without explicit place or location mentions, except in the names of the participants or the residents of the area, e.g. "Tibetan protesters involved in rioting last year". Since a riot is associated with a locality by definition, the participant mentions containing place names will be regarded as location information in and of themselves. That is, these events will be labeled as protest events despite the location of the event is not explicitly mentioned.

xiv. News articles compiled from any newspaper will contain reports about countries other than the EMW project's case countries. In this level of annotation, we do not exclude news reports about countries other than our cases. As long as the protest event mentioned in the article conforms with the definition and criteria laid out in this manual, it will be labeled as a protest event.

## 2.1.5 Examples of cases that should not be considered protest events

In order to draw a more complete picture, this section will provide several illustrative cases which are not included in our event definition. Note, again, that this is not an exhaustive list but a collection of cases where it is apt to mention specifically that the events we encounter in the news articles are outside our scope.

An event is excluded from annotation either if it is outside the scope of our event ontology of contentious politics or its reporting in the news article lacks the necessary event characteristics that are defined above.

### *2.1.5.1 Events that fall outside our protest event definition*

   i. Events that cannot be considered political protest or mobilization are, obviously, not included. Petty crimes, fights or clashes between civilians that lack political, economic, or identity-based motivations, or actions targeting officials or political figures simply for personal reasons are thus outside our scope. At times certain cases of violent crimes might involve actions that can readily be associated with political violence, such as using bombs to rob banks or pirate attacks. Care must be taken to make sure such events have political motivations.

   ii. We do not consider individual public actions such as petitions or litigations of public character that lack collective support and remain on the individual level as protest events.

   iii. Events that are initiated by the state or government actors that are not actively participated by civilians are outside our scope. These might be military operations on non-state organizations or other states, acts of repression on members of the public (note that the latter's active resistance would count as an event), official ceremonies, celebrations, commemorations organized by state/government authorities, events that take place within the auspices of the parliament and would be part of common government procedures (e.g. politicians leaving the parliament, having heated arguments, obstructing decision making, etc.).

   iv. Protest actions by members of legislatures during meeting sessions are mostly habitual events and are associated with the routine functioning of these bodies. Acts such as disrupting speeches or groups of members disrupting sessions, verbal protests, banging on desks, rushing towards speakers, walking out of meetings, even fistfights, etc. are all usual events in parliaments that have little to no consequence outside the legislatures. Such events are not included in our protest event definition.

   v. Events that are part of election campaigns and, unlike election rallies, do not mobilize civilians are not considered protest events. Examples would be canvassing, distribution of election brochures, leaflets, etc. that are not actively participated by members of the public.

   vi. Simply declaring certain demands and grievances, as in submitting memorandums, giving declarations to the press, without the said act of declaration itself becoming a public spectacle is not considered an event. However, a press declaration outdoors is a public spectacle and a prominent way to protest in certain countries and as such is included in our event definition.

   vii. Sometimes the news stories are about general contexts or situations of conflict between a set of actors that are very prone to cause or likely to contain protest events but are not events themselves. Discerning events from situations might be trickier than it seems at first sight as the latter might be so concrete and specific as to be given concrete times and places. Annotators must be careful with expressions such as tension, conflict, disagreement, enmity, etc. as these do not denote actions (events) but rather situations. If the article does not include a concrete confrontation event (a clash for instance) that are caused by the conflict, it should not be labeled as a protest article. Examples of conflictual situations which do not count as events in and of themselves include the tense atmosphere of prolonged conflict between a state and a militant organization which is strained in summer months when the state engages in military operations, the conflict between the managers and workers of a subcontracting firm over unpaid wages, the distress created among the workers of a shipyard by recurrent job-related accidents, the tension between two religious communities living in the same town before a prominent religious holiday, etc. The protest events we would observe in such situations, and thus label as such, would rather look like: militants' retaliation attack on military outposts, workers beating up their bosses, the shipyard workers occupying and stopping the work in the workplace, the mobs belonging to each community ransacking each other's small businesses, etc.

viii. Similar to the previous case, some news stories contain statements that give information about the background or the context of one or more of the story elements. Such statements might contain general allusions to, or summaries of multiple events. Such events might be very likely to be contained in our event definition. In such cases generally speaking the annotator must be careful to mark events that might elude a superficial reading. However, we are after concrete events that we can identify rather than summaries of event-rich contexts. Extra care should be taken to mark only the events which are specific, concrete, and can be singled out. Distinguishing concrete events from eventful contexts can be tricky as the difference is hard to define comprehensively and the clues in the text can be subtle. Consider the difference between the sentences below:

    ○ The city is a far cry from the hotbed of riots that it was until the beginning of the decade.

    ○ The city is a far cry from the hotbed of riots that it was during the last months of 2009.

    The second sentence is very likely to be referring to a specific series of riots while the first sentence implies that the city used to have a certain characteristic that made it more prone to be the scene of riots. The second sentence would make the article containing it a protest event article.

ix. Events that are initiated by foreign governments, armies, or any institution of a foreign state are outside our scope.

x. Cases of military coups are not included in our protest definition. Cases of rebellion, mutiny, and civil war that are instigated by state officials (e.g. part or whole of the military) are non-civilian actions and thus go beyond contentious politics. The annotators should consult the domain experts about cases of civil wars where they find it difficult to decide.

- xi. Events of mass migration have been salient issues all over the globe, particularly in the Middle East, Europe, and the Americas, in the last decades. Passage of masses of migration caravans through cities on their trajectory, camping at and attempting to cross international borders, and residing collectively in refugee camps might resemble at times contentious political events. Even though migration due to reasons of poverty, human rights abuses, or war has aspects that challenge state authorities in designation and passage countries, we do not treat it as a contentious political event on its own. However, migrating masses tend to engage in contentious political actions that we do include in our event ontology. We include actions such as marches, slogan shouting, carrying banners/posters, clashing with security forces, and group clashes that involve migrants in host countries as contentious politics events, but try to exclude migration itself while annotating.

### 2.1.5.2 Events that are included in the event definition but lack necessary characteristics

- i. As stated above, the event described in the article must have taken place or be an ongoing event to be annotated. There are some cases in which the event in question is/was expected to take place but did not. These cases are outside our scope and will not be annotated.

    a) The announcement or mention of events that are to take place in the future and hence reported in the future or conditional tenses are not to be included.

    b) Sometimes the articles may mention events that are supposed to take place the same day the article is published, and hence report them with a different, stronger sense of certainty compared to what would characterize the news about future events. In these cases also, the norm in the news reports is to use future or conditional tenses, i.e. whether and how the events "will" or "would" take place. That is to say, the clue would again be in the tense used in the narrative.

    c) Instances of vowing (an earnest promise to perform a specified act or behave in a certain manner) or threatening to carry out certain actions that are not carried out. The exception here is the situation mentioned in the 8. case in the "Some event cases that are included" section whereby the threat of the action itself has the effect of a protest, i.e. bomb threats, death threats, and self-immolation or other suicide threats which also frequently become spectacles themselves.

    d) Planned, threatened or intended actions that are thwarted or disrupted before happening such as a strike or demonstration that is disallowed by authorities, or apprehension of militants that are planning to carry out an action are not considered events. The exception to this are cases of threats and attempts of violence, as it was mentioned in the previous section.

    e) Cases of bomb hoaxes and similar deceptions are not to be considered protest events.

f)  Sometimes a news article will be about judicial procedures such as prosecution or court cases about protest events. Care should be taken when deciding in these cases so that the articles which only contain protest mentions which do not refer to the actual events but legal or procedural concepts are labeled as non-protest articles. An article in which the ***only*** protest mention is e.g. "cases of rioting and murder are registered against them", is not a protest news article.

ii. The second necessary aspect of the events to be annotated concerns their spatial characteristics.

a)  As it was mentioned in the necessary characteristics section, vague time expressions which nevertheless inform us that an event is a current affair are acceptable, but event location info is crucial to make the event mention sufficiently concrete. The events that do not get specific location information attached are outside our scope. In cases where event place information (i.e. a specific geographical location more specific than the country) does not exist, a facility (building, campus, factory, etc.) provides the necessary concrete spatial dimension. Event mentions or summaries which do not come with any location information in terms either of geographical place or facility within the article are not to be annotated.

b)  Events that gained historical significance and/or have become so iconic that they are mentioned with their specific name are a frequently encountered case of lack of location info. Gezi Events, 9/11 Attacks, Phoolan Devi Assassination are prominent examples. When these and similar events are not current and mentioned only due to their historical significance, it is usually straightforward to exclude them. But when they are relatively recent, it would be harder to decide whether to include them or not. The location rule is a more certain guide in deciding in such cases.

c)  Another case that might be difficult to discern whether to include or not is when the news article contains a story about the survivors, victims, or people affected by a certain event. In these cases, too, the articles are not annotated if the location information of the event whose victims are the focus of the story is not provided.

# 3 Sentence Level Protest Annotation Manual

This task aims at identifying and labeling sentence(s) that contain protest events in the news articles. It follows the document-level protest labeling task which identifies news articles that contain protest events as defined in the DOLPAM. Once the news reports are classified as containing protest events, what remains is to identify where in the article the relevant event information is presented. In determining this location, the linguistic unit we evaluate is the sentence. The sentences which contain event information will be labeled as event sentences. Those that do not contain event-related information will be labeled as non-event sentences. Finally, those sentences which contain mentions of events that have not happened, i.e. future events, will be labeled as planned event sentences. Read each sentence of the protest documents in the annotation tool and label it accordingly.

## 3.1 Classifying Sentences

As an abstract procedural framework, the task can be divided into two phases. The first is to determine whether the sentence contains any event reference, thus making it eligible to be labeled as either 1 or 2. Once the non-event sentences are labeled as 0 and thus excluded, what remains is to decide whether the event took place in the past, is an ongoing event or has not taken place. The past and ongoing events are labeled as 1, the rest 2.

### 3.1.1 Event sentences

Event sentences must contain an expression that refers directly to any protest event that makes the document eligible for being classified as a protest article. Such reference can be any word or phrase which denotes the said event. They can be direct expressions of the event or the pronouns which stand for the event (e.g. protest, march, demonstration, strike, clash, incident, event, etc.). These expressions are labeled as event triggers in the next annotation task. Thus, the detailed definition of event triggers is in the TOLPAM below so please refer to it for a better grasp of the concept. The event sentences must unambiguously indicate that the events in them have happened in the past or are ongoing events. For additional information about this principle and our event definition, please refer to the DOLPAM.

### 3.1.2 Non-event sentences

Non-event sentences are the ones that do not contain any event references past, present, and future. Note that they might contain event-related information but this is not enough in and of itself to label them as event sentences insofar as they do not contain a direct reference to the event itself.

### 3.1.3 Planned event sentences

Planned event sentences are those sentences that contain references to events that have been planned, announced, threatened, or attempted by a given actor but have not taken place for one reason or another. They can be events that are expected to happen in the future, events that had been planned in the past but have not taken place, or events that any social agent threatens or declares to carry out. As it was mentioned in the previous manual, threats, and attempts of violence are to be treated as protest events so they will be labeled as protest event sentences rather than planned.

## 3.2 Notes and examples

i. The event sentences must contain at least one expression which denotes the event. These expressions can be:

   a) Expressions that denote an action that is the whole or part of a protest event e.g. clash, strike, demonstrate, protest, attack, hit, blockade, shout slogans, sit-in, kidnap, assassinate, etc.

   b) Generic or dummy expressions which stand for the event e.g. incident, event, etc.

   c) Pronouns and demonstratives (such as "it" and "this") that refer to events.

ii. The expressions listed below are not to be considered event expressions as defined above, that is, they do not make a sentence an event sentence by themselves:

   a) Expressions which denote the participants or other actors such as protesters, agitators, demonstrators, attackers, etc.

   b) Expressions which denote situations that cause or contain events, but neither are events themselves nor stand for events e.g. situation, tension, conflict, etc. (Please refer to the DOLPAM for a detailed explanation of the difference between events and situations.)

   c) Expressions which denote the consequences of events, e.g. death, injury, damage, captivity, etc.

   d) The word "protest" when it means the feeling or attitude of objection e.g. "There were murmurs of protest among the committee members after the meeting was postponed for the second time."

iii. The sentences which refer to the protest events in an abstract or general sense are not event sentences. e.g. "Police forces were heavily deployed in the area to deter any such agitation" or "The inhabitants of the city are no strangers to violent clashes."

iv. The necessary characteristics of protest events such as the presence of civilian actors, time and location necessities hold for event sentences as well, however, we do not expect the sentences themselves to satisfy them. If the event that is mentioned in the sentence satisfies the criteria in the overall document, the sentence is to be labeled 1. On the other hand, if the event mention in the sentence conforms to the criteria neither in the sentence nor in the document, the sentence containing it will not be labeled 1.

v. Be mindful of the sentences which contain event references but mean or imply overall that the said events did not take place, or were not protest events. These sentences should be labeled [0]. e.g. "The governor criticized the press for covering a petty street fight as a political clash." "The aid convoys are advised to use those border crossings where there are no clashes and disruptions."

vi. Planned event references must be made to concrete possibilities. In other words, there must be a threat, an announcement, or an attempt at an event. Perceptions of risks or threats are not to be labeled as planned events. e.g. "With the union representatives voicing their dissatisfaction with the negotiations, the possibility of a transportation strike looms large." "The embassies have warned their personnel to avoid crowded places due to the risk of a terrorist attack." "'We believe the organization was planning a major strike during the upcoming celebrations', another police official said."

vii. Similar to the above point, the threats voiced by a given actor must be specific enough to warrant being labeled a planned event. Some threats to take an unspecified action might not involve protest events. e.g. "The party spokesperson threatened dire consequences if the ruling coalition goes forward with the motion." "Their leader vowed that they would retaliate in kind to any such action."

viii. As explained in more detail in DOLPAM, remember that the threats or attempts of violent actions such as assassination, bombing, and self-immolation are to be considered events themselves, and therefore the sentences that contain them will be labeled 1.

ix. The speeches given during a rally or a protest gathering are not protest mentions themselves.

x. As in the previous document-level annotation task, reports of events occurring in countries different than the source country will be annotated on this level.

xi. While annotating news reports about migration events, take care to remember the explanation on the topic in the DOLPAM section, and try to distinguish contentious politics events that fall in our scope which occur during mass migration from the act of migration itself. Sentences, which contain expressions like "caravans marching", should be considered carefully to determine whether the march in question is the movement of migrants or an actual protest march. There will be cases in which the two are intermingled and therefore not readily distinguishable. In such ambiguous cases, we prefer to err on the side of exclusion and not consider such actions protest events when there is no explicit mention of a protest action.

xii. As it was mentioned in the DOLPAM, event references which are legal concepts and mentioned in the context of judicial processes about actual events are not to be considered protest events themselves and will not be considered event sentences. e.g. "The activists were charged with rioting, obstructing the police and damage to public property." In this sentence, the protest event mentions do not refer to actual events but their legal concepts, hence this is not an event sentence.

xiii. As it was mentioned in section 2.1.5.1. of DOLPAM as well, the valid event references which we label as protest events must be to actual events rather than general, contextual qualities of other story elements. Consider the example below.

> "The organization has been known to have been involved in bombings in Mumbai."

> "The organization has been known to have been involved in the bombings that took place in Mumbai."

The second sentence contains event information that makes which makes it an event sentence as "the bombings that took place in Mumbai" refers to specific bombing events. On the contrary, in the phrase "bombings in Mumbai" in the second sentence, such a specific reference is not present.

# 4  Token Level Protest Annotation Manual

## 4.1   Introduction

Linguistic annotation of the protest event-related information in the news reports involves the task of labeling linguistic units (words or phrases) that gives such information within the text of the report. The third main task of the GLOCON GSC annotation process is this task of labeling the tokens comprising sub-sentence linguistic units that appear in the text as protest event-related information. The documents that will be annotated on this level have already been annotated on the document and sentence levels, and thus are known to contain protest events. The task should start with the detection and labeling of event references (see terminology), which act as anchors of the whole annotation process on this level. Alongside event references, other words or phrases that include values or entities that give event-related information such as its time, place, participants, organizers, targets, and the like will be enumerated below. These entities or values will be referred to as event arguments.  Each information type and the tag(s) designated to them for labeling will be described briefly. General rules about linguistic information extraction and specific rules about each tag will be provided in detail with many examples in a variety of languages and from countries within the scope of the GLOCON database.

This manual aims to assure consistency and clarity in the annotation process. Since it contains a lot of detailed information and rules about a multitude of variables, please keep in mind to come back to it for reference throughout all your annotation effort. While it is advisable to read the manual to get a general sense of the specifics of the annotation process, it is intended for regular and repeated reference; so do not feel obliged to learn or memorize all the rules contained in it. It is built on the event definition (or ontology) developed in the Document Level Protest Annotation Manual (DOLPAM). You should be familiar with this ontology to be able to recognize the event-related expressions in the documents; so please refer to it before carrying on with this manual.

The labeling process is performed in the "Annotation editor" of the FLAT (Folia Linguistic Annotation Tool) and consists of five foci; event, participant, target, organizer, and semantic classes. These are found under "Annotation Focus" in FLAT. The document opens by default in the "Annotation editor" and "Event" focus. In the annotation manual, the tag names are written in lower case letters and multi-word tags are separated with an underscore. Read the rules below carefully and report any issue you may observe or any suggestion you may have to the advisors.

## 4.2   Terminology

- **Event**: Events that fall in the scope of contentious politics. Refer to DOLPAM for a detailed description of what we consider a protest event.
- **Event reference:** Any word or phrase that expresses an in-scope event in the text. Two types of event references are defined in this manual. "Event types" are first occurring event references that express the specific nature of the events, whereas "event mentions" are any subsequent words or phrases which express and/or stand for the event, including pronouns and demonstratives. We have a particular event definition (See the DOLPAM for further details on this definition) and only the news articles that conform to this definition should be annotated. If you think the news article does not contain an event as defined in the DOLPAM, do not do any annotation on that news article. Simply mark the "Event" field in the Metadata Editor section of the FLAT, as "No" and notify the advisors.

- **Event argument:** Entities or values that are connected to protest event references and give information about them in event sentences.

- **Event sentence:** In the news article, "event sentences" are all the sentences that contain the event_type and event_mention information (Please see "Type of Event" section for further information about event_type and event mention). They are generally located in the first paragraph of the news article and usually contain vital information such as event time, place, main participants, etc.

- **FLAT-sentence:** The annotation environment may represent text fragments as partial sentences, single sentences, or several sentences in the same line, which is given a number on the environment. These lines are referred to as FLAT-sentence.

## 4.3 General Rules of Token Level Annotation

### 4.3.1 How annotation proceeds

First, read the whole of the news article you are going to annotate before starting the annotation. If the tagged expression occurs multiple times in the article, keep in mind that our priority is the information directly related to the event(s) in the article. We aim at annotating the most relevant entities to the event. This information is usually provided in the document title and/or in the event sentence(s) (see the terminology section for the definition of event sentence"). Therefore, please follow the steps below to avoid unnecessary tagging or missing any relevant information:

a) Identify the title of the news article and annotate all event-related information in the title.

b) Identify the event sentence(s) and annotate all event-related information in those sentences. In cases where the tagged word is used more than once in the event sentence(s) then all those words should be tagged. E.g. "the problem arose in the village after a TDP supporter won the third ward of the village in the panchayat election held on August 14." Both occurrences of "village" in the sentence should be tagged as rural_location_identifier (See the relevant section below for further details on how to use rural_location_identifier tags).

c) There will be no annotation outside the event sentences, that is, sentences that do not contain an event_type or event_mention tag will not be annotated.

Observe the following sentences as an example of this principle:

- "Samajwadi Party workers also held a demonstration at Jantar Mantar on Thursday morning demanding resignation of up chief minister Rajnath Singh and union home minister L. K. Advani for their negligence which allowed the assassination.
- "SP general secretary Usha Yadav said the party leaders also visited Phoolan Devi's residence at Ashoka road to express condolences to the bereaved family."

In this example, only the first sentence is event-related. The entities which are related to the protest (in the first sentence) such as SP, are not tagged in the second sentence because they occur in an irrelevant context.

d) In case a tag is not used for an article, we will assume that the relevant information was not in the article. There is no way to retrieve the information lost due to missing labels. When you hesitate to use a tag, erring on the side of inclusion, that is, putting an unnecessary label may be preferable to losing information. In such cases, you can give a lower confidence value to your annotation, and add a comment which explains your doubts. You will find the confidence level bar once you check the "confidence" box in the annotation window which pops up when you select a phrase to annotate. Please try to give a brief explanation for the low confidence by adding a comment to avoid future confusion.

## 4.3.2   When to tag event arguments?

All event arguments listed in this manual will be labeled when they are syntactically and/or semantically connected with the event in a non-negative way. This principle can be elaborated in four ways:

i. The immediate and most significant implication of this principle is that we will not code sentences that do not contain event references. Event information will only be tagged when they appear in event sentences, that is when they appear as event arguments.

ii. An entity or value must be connected to the event in the event sentence to be tagged as an argument. For instance, a time value or place name that an event sentence contains must be given as the time or place of the event. A direct syntactic connection of this sort is usually straightforward. E.g. "The protesters gathered at the plaza in the early afternoon". An indirect and semantic connection can also make certain entities and values event arguments. E.g. "Tuesday started as an ordinary weekday for thousands of residents, who were unaware of the chaos the afternoon's attack would spark". Here, even though there is no direct syntactic connection between the time value "Tuesday" and the event reference "attack", the sentence establishes the semantic connection, making the former the time argument of the latter.  That said, the connection between the event and argument must be unambiguous. Care should be taken not to infer relationships between events and entities that complex sentences might include but not connect. E.g. "A Delhi court found him guilty of perpetrating the attack that cost the lives of innocent civilians". In this sentence there is no indication that the attack took place in "Delhi" and therefore it should not be tagged as an event place.

iii. The modality of certainty will not play a role in showing the relationship between events and arguments. This means that we annotate arguments even when the documents use qualifiers such as supposed(ly) or alleged(ly) before them. E.g. "the police have reportedly rounded up 11 members of the BKU for allegedly attacking one Kishan Singh" In this case even though the actor status of "11 members of the BKU" for the attack is "alleged", we treat them as true event arguments and label accordingly.

iv. Unlike the case of uncertainty modalities mentioned above, the even-argument connection will not be assumed in negative sentences. In sentences that establish a negative connection between events and arguments, we will not label negated entities as arguments e.g. "BKU leadership denied the organization's involvement in the attack". This sentence establishes a negative connection between BKU and the "attack" event. Thus, BKU is not taggable as the organizing actor of this event.

### 4.3.3  Overlapping Annotation

Sometimes certain expressions or parts of them give pieces of information that correspond to more than one tag. Overlapping annotation refers to tagging the same expression (or a part of the expression) with more than one tag. In the rules enumerated below, you will find the cases where overlapping annotation is required as well as cases where it is not allowed. Note that the rules are exhaustive, that is, overlapping annotation will be allowed for only the cases mentioned below.

i. Tags on all kinds of event information in the document title will overlap with the document_title tag.

ii. Participant type expressions that also relay information about other participant characteristics such as ideology, religion, ethnicity, caste, and SES will be tagged with all applicable tags. e.g. "Teachers are determined to continue their strike unless their demands are met." [Teachers]=participant_type, participant_SES e.g. "The city witnessed a massive rally of thousands of angry Maoists today" [Maoists]=participant_type, participant_ideology

iii. Organizer type or name expressions that contain information annotated with other tags in the organizer focus, i.e. organizer, ideology, religion, SES, caste, and ethnicity. e.g. "Over 1.000 people were present in the Communist Party of India (Marxist) rally today" [Communist Party of India (Marxist)]=organizer_name, [Communist]=organizer_ideology, [Marxist]=Organizer_ideology

iv. Named entity expressions will only be annotated with "name" tags. There will be no "type" tags overlapping with facility, organizer, and participant name expressions.     e.g. [Safdarjung Hospital]=facility_name (i.e. "Hospital" will not be tagged with facility_type) e.g. [Karnataka State Government Employees Association]=organizer_name (i.e. no other tag will be used in this expression.)

v. Arguments that belong to multiple events (that exist in the same sentence) and give different types of information about each of them will have all tags necessary. For instance, if the participants of event 1 are targets of event 2, overlapping annotation will be allowed for participant and target tags with their respective event number comments.

vi. In situations where the facility of the protest action (i.e. the building or built environment which serves as its location) is at the same time its target (i.e. the entity against which the protest is taking place), both tags should be put accordingly. This situation usually occurs when the facility expression comprises premises of the entity (organization, government, etc.) that the participants of the event protest against. Tag span rules of facility and target tags (see below their respective sections) will determine the tokens that are shared by two tags. For instance, the facility tags require their preceding prepositions to be included in the tag span, whereas target tags will never include prepositions. Hence, in the phrase "protesters pelted stones at the Office of Traffic Monitoring," since the said "Office" is the named entity which is the antagonist of the protesters, it should have the target_name tag. Also, since the expression "at the Office of Traffic Monitoring" comprises the facility information, the whole expression will be tagged with the facility_name tag. In the final picture, the phrase "Office of Traffic Monitoring" will carry two overlapping tags, facility_name and target_name.

### 4.3.4 Tag span rules:

#### 4.3.4.1 Tag span continuity

Even though the FLAT annotation environment allows non-continuous tag spans, all tag spans which cover any kind of event, entity or value must be continuous. In other words, there should be no tokens (including punctuation) that are untagged between the tokens of a tagged expression.

While annotating an expression that contains multiple tokens, make sure the tag span does not exceed the end of a FLAT sentence. You can recognize these sentences as numbered on the left side of the document like rows in a table. If the phrase overflows to the next sentence, annotate the part of the phrase in the second sentence with a new tag, if that sentence is also an event sentence.

#### 4.3.4.2 Tag spans in complex expressions:

As a rule of thumb, we must aim to include in the tag span as few words as possible. The more words are in a tag span, the less the likelihood of inter-annotator agreement, and thereby consistency. For many tags, coding single words for event references, entities and values is ideal. Adjective, adverbial, noun, or prepositional qualifiers of event and argument entities are left out of the tag spans most of the time (exceptions are in tag descriptions below). Yet there will be instances where a single word will not indicate an entity in a meaningful way, necessitating a wider tag span. While tagging such complex expressions, care should be taken to identify the smallest meaningful phrase which captures the entity in a meaningful way. The most general exception to the one-word tag span principle concerns compounds. Compounds are single units of meaning that are formed by combining more than one word such as "someone", "likewise", "camcorder", etc. Some compounds, such as "floppy disk" and "ice cream", are written separately and look and act like noun or adjective phrases. These might be difficult to discern from phrases, which entail descriptions of qualified entities and are thus not single units of meaning as a whole. Consider "floppy disk" which designates a single entity, and is not a description (i.e. it does not mean "a disk that is floppy"). Common examples of compounds that we encounter often in protest annotation are phrasal verbs such as "flare up" and "walk out" or expressions such as "hunger strike", "take to the streets", "toddy tapping," and the like. Since compounds are single units of meaning, i.e. designate single entities, when they are tagged, the tag spans must cover all the words that make the compound (i.e. as if they were written in a conjoined way). On the contrary, modifier words in phrases must be left out of the tag span as they do not change the nature of the concept that is tagged but merely describe it (e.g. school teachers, landmine blast, etc.). To distinguish compounds from ordinary phrases you can keep in mind that the individual words which form them usually lose their original meanings once they are in a compound, that is, the compound means something different from each of its word parts. You can also remember that they usually have their own definitions in dictionaries, unlike noun or adjective phrases. Below, under tag definitions, you will find tag span rules spelled out for certain complex expressions that require special treatment.

### 4.3.4.3 Tagging punctuation and articles

The punctuation marks, including quotation marks and parentheses, at the beginning and end of the expressions that are tagged, should not be annotated. This principle does not apply to punctuation marks in the middle of the expressions tagged.

Indefinite articles (a, an) are not to be included in the tag when they occur before or after the expressions to be tagged. Definite articles ("the") will only be tagged if it is part of the official name of an entity, e.g. "The Supreme Court". Lowercase "the" will not be tagged.

## 4.4 Document Information Tags

## 4.4.1 Title of the document

The news article will have a title by default most probably at the top of the page. Titles can usually be distinguished by their linguistic characteristics—for instance, they are usually not complete sentences. Also, they tend to precede the publishing time and place expressions, which are easier to distinguish. You should mark the document titles with the document_title tag.

**Tag rules**
  i. If the document title is on top of the news article, mark it. This is the case most of the time.
  ii. Some articles may contain more than one news report independent of each other. Some of these reports may not include any protest events, in which case their titles will not be tagged. Mark only titles of the articles that include at least one event. You should insert a note in the comments section of the tag for the second and following titles in this case. The comments should start with "Event 2" for the second title and go on Event 3... if there are more titled reports.

iii. In multiple report cases, the titles that follow the first one might not be as easily discernible as the first title, which is usually on the top. The flat environment does not contain font size or color variety that distinguishes the titles from the text body. Try to single out the titles from punctuation, inconsistent upper or lower case use, etc.

iv. We use this tag only to identify the news article titles as they were headlined by the newspapers. In case there is no title at the top of the article, do not search for it in between the lines. Simply notify the advisors about the missing title.

v. The title generally contains key information about the event. You should first annotate the words or phrases that contain event information with relevant tags, and then annotate the entire title with the document_title tag.

### 4.4.2 Document publication time

The publication date of the news article will be provided, commonly in the first flat sentence, as it is extracted from the HTML file of the news article. Please mark it with the event_time_published tag. Make sure that you mark the given expression in full, including the date, time and time zone -- if provided. Don't forget to mark the punctuation marks (commas) within the time expression. e.g. "Sep 8, 2001, 23:34 IST"

### 4.4.3 Document publication place

The publication place of the article is also extracted from the HTML file and will commonly be located on the first flat sentence, near the document title and publication time. Publication places can be distinguished due to the way they are written as stand-alone place names and do not form part of sentences or other phrases. The event_place_published tag will be used to label the publication place. In cases where there is more than one publication place, e.g. when the state is mentioned as well as the city, each place name must have an event_place_published tag.

## 4.5 Event Reference Tags

### 4.5.1 Type of the Event

The event has a central role in our annotation efforts. You should first detect the event in a news article before you do any annotation. Events that can be included in the broad term contentious politics are in our scope (for the sake of simplicity, we use protest events to refer to contentious politics). Contentious politics refers to any form of grassroots political action, and actions of political and non-governmental organizations that are aimed at mobilizing the public in the name of political demands and grievances. Such actions are outside the institutional and legal forms of political participation such as voting, litigation, etc. For details as to our event definition and various event types, refer to the DOLPAM.

Examples of protest events can be given as follows:

Demonstrations, rallies, marches, press declarations that take place outdoors in public, gatherings, meetings (when referring to public rally-like meetings), sit-ins, unofficial commemorations with mass participation (e.g. funerals or burial site visits), acts of civil disobedience, and nonviolent resistance, collective petitions, collecting signatures, strikes and any other work-related protest events (slow-downs pickets, etc.), hunger strikes, death fasts, lynchings, lynch attempts, kidnappings, occupations, boycotts, cyber-attacks, hacktivism, hijacking, self-immolation, armed and unarmed violent attacks (by non -state actors), armed militancy actions against state security forces, unarmed clashes with security forces, clashes among groups of civilians caused by political issues, and the like.

Event types that are not among these examples should be evaluated based on their relation to definitions provided in the DOLPAM. Any doubt should be communicated to the experts.

The protest events that we annotate must be current or past events. Events that are announced, threatened, or expected to take place in the future in any other way will not be annotated. The only exception to this is when a threat of violence is issued by an actor, or a violent action attempt is thwarted, e.g. a death threat or a planted bomb that is diffused before going off. Self-harm threats and attempts are to be considered violence attempts/threats, e.g. a self-immolation attempt or threat. Such threats or attempts are thought to have the effect of events carried to completion and are thus regarded as events themselves. General or conceptual mentions of events, instead of specific happenings, will not be annotated either.

The type of the event should be annotated as per in the examples below:

- Thousands of people **rioted** in Port-au-Prince, Haiti over the weekend.
- The union began its **strike** on Monday.
- Protesters **rallied** on the White House lawn.
- The **rioting** crowd **broke windows** and **overturned cars**.
- A crowd of 1 million **demonstrated** Saturday in the capital, San'a, protesting against Israel, the United States and Arab leaders regarded as too soft on Israel.
- In Ramallah, around 500 people **took to the town's streets**, **chanting slogans** denouncing the summit and calling on Palestinian Authority leader Yasser Arafat not to take part in it.
- For weeks Italian Jewish groups, World War II veterans and leftist political parties have staged **protests** against a meeting between the pope and Haider, arguing that a papal encounter would lend the Austrian politician legitimacy.
- More than 40,000 workers were back at their jobs Thursday following a 1-day **walkout** that closed social welfare offices and crippled public medical services. During the work **stoppage**

Wednesday, local residents were unable to register marriages or get documents for real estate transactions.

- Shah's supporters also had gone on the **rampage** outside the principal's office on Tuesday.

**Tag rules**

i. Tag span rules:

a) Adjective, adverbial, prepositional, or noun qualifiers will not be included in the span. e.g. landmine ==blast==, token ==strike==, forcibly ==stopped==, communal ==flare-up==, indefinite ==hunger strike==, fierce ==encounters==, twin militant ==attacks==, protest ==demonstration==, protest ==march==.

b) In verb event types, only the verb will be in the span. e.g. ==boycotted== a meeting, ==rioting== with weapons, forcibly ==stopped== the auto, fight it with ==bombs==, landmine ==exploded==, hit by crude ==bombs==, ==refused to work== extra hours.

c) Auxiliary verbs will not be included, except when the verbs are in the negative form e.g. have ==blocked off==, will ==protest==, are ==demonstrating==, ==did not start work==,

d) In noun event types, only the noun will be in the span e.g. laid an ==ambush==, forced a ==bandh==.

e) If the verb or noun itself is not meaningful or is dubious as a protest action, include the minimum required words. Since tag spans cannot be broken, this might necessitate the inclusion of non-event type words. e.g. ==Violent reaction==, ==violent culmination==, ==took to the town's streets==, ==did not start work==, ==refused to work==, ==shouted slogans==, ==written slogans==, ==hung posters==, ==carried banners==, ==hurled stones==.

ii. If the article contains multiple events, mark all. If there are multiple events in a news story, the events that have different times, places, facilities, and participants from other events should be annotated separately. The rules about event separation are a bit complex so please refer to the Event Separation section in this manual.

iii. Demonstrations, gatherings, electoral rallies, etc. organized by political parties, organizations, or representatives of political parties should be included. However, events happening under the auspices of the parliament such as politicians leaving parliamentary sessions, disrupting them, or having heated arguments with each other are accepted as usual proceedings within parliaments and thereby not considered as events. But, if politicians engage in an unusual protest action that can be considered outside of the formal parliamentary procedures and proceedings, (e.g. such as hunger strikes, sit-ins) these will be considered protest events.

iv. Spontaneous gatherings or demonstrations with no organizer or leader should be annotated as events.

v. Be aware that Hindu rituals such as idol immersions and akhand paths are not protest events in themselves. However, they can become means or scenes of protest events. In an expression like "Clashes broke out after idol immersion", only "clashes" should be annotated as event type but in a sentence like "An akhand path was organized to protest government's decision", this ritual is deployed as a protest itself.

vi. Public interest litigations and writ petitions are to be tagged as event types in case they have political motivations, and are participated by multiple members of the public.

vii. Expressions such as "kill," "injure," etc will be not tagged when they refer to consequences of violent events, rather than the act of killing or injuring, intentionally. Bear in mind that in English "Three people were killed in the attack" is identical in meaning to "Three people died in the attack" so in both sentences only "attack" is the event type.

viii. The word "protest" is a very versatile word that can sometimes mean being in or declaring opposition to something, or a strongly worded refusal or rebuttal. Be mindful about such situations where the word protest does not stand for events that are in our scope.

ix. Similarly, the word protest will not be tagged when it means intention or aim rather than the act of protest itself, as in expressions such as "They held a rally in protest against the management". In a sentence like "They shouted slogans, protesting against the tariffs", only "shouted slogans" will be tagged as the event type.

### 4.5.2 Event Mention

Any word that refers to the event type which occurs after the first occurrence of the event_type will be tagged with the event_mention tag. Pronouns and token words which refer to the event type are included in the tag's scope. Token event words are generic expressions that denote the events (i.e. words like incident, event, protest, and agitation).

**Tag rules**

i. For every separate event, there will only be one event_type tag which will be the first occurrence of the event reference, provided it is not a token event word (see below). All the following event references and pronouns that stand for the event type will be tagged event_mention.

ii. Event words in the document title will be tagged with the event_type tag. The first occurrence of the event word outside the document title will also be tagged with the event_type tag.

iii. Token event expressions will be tagged with the event_mention tag even if they are the first occurring references to events (see the examples below). Token event words are semantically indeterminate, that is, they can designate various types of events, while true event type words are more descriptive, and restrictive in meaning. It is possible to generalize the distinction such that the token event words can stand for multiple categories of events defined in the DOLPAM, namely, demonstrations, industrial actions, group clashes, and political violence and militancy. The words like incident, event, protest, and agitation can stand for events from all (or more than one) of these categories. While restrictive true event type words do not have this variety (e.g. a hunger strike can only be classified as a demonstration.) Token event expressions will always be tagged event_mention. The only exception to this is the case when token words are the only event words that are used in the article to refer to the events. In such a case, the general rule will hold and the first occurrence of the token word will be tagged event_type and the following references will be tagged event_mention.

iv. Event separation logic will be the same as any other tag. That is, any event mention tag that is used for each separate event will be numbered in a comment.

v. The sentences which contain the event_mention will become event sentences, thus any relevant event information in these sentences will be tagged even if the event_type tag occurs in other sentences.

### Examples

Observe the following examples where the event_mention tags are highlighted green and event_type's are in yellow.

CPI(M) stages protest ==rally== in Bhavnagar. The Bhavnagar unit of communist party of India CPI(m) on Friday staged a **demonstration** opposite the local post office here...

–The word "rally", the first occurrence of the event word, is tagged event_type, demonstration which refers to this event, is to be tagged with event_mention instead of event_type.

Commenting on the ==strike== which was flagged off on Monday, the union secretary stated "**it** will continue as long as our demands are not met."

–The word "it" stands for the event_type "strike," and thus, is tagged with event_mention.

In a separate **incident** in the neighboring area of Awantipore, an air-force vehicle was damaged and two security force personnel were injured when a landmine ==exploded==.

–Here, the truly descriptive event word is "exploded," which is tagged event_type, incident is indeterminate in comparison, and thus, will be tagged event_mention, even though it occurs later.

The students organized a **protest** by ==marching== against the payment seat decision.

–Here the token event word "protest" is tagged event_mention as any contentious politics event can be referred to as protest, whereas the word "march" is tagged event_type as it directly indicates what the protest is, and thus is a true event type.

A variety of NGO's expressed their support for the ==agitation== of teachers in the capital.

–Considering this sentence is the only one which mentions the event in the article, the token word agitation is tagged event_type.

### 4.5.3 Semantic category of events

The "Semantic" focus in the flat environment contains the tags to be used when labeling the event triggers (types or mentions) according to the protest type that they indicate. The semantic tags will be placed on each event trigger in an overlapping manner. That is to say, the tag spans of the semantic category tags will be coterminous with the tag spans of event type and event mention tags. The tags and their definitions are as follows:

#### *4.5.3.1 Demonstrations*

A demonstration is defined as a form of political action in which a demand and/or grievance is raised outside the given institutionalized forms of political participation in a country. The event aims to draw the attention of politicians and/or the general public to said goal by making it visible in the public sphere which implies any space which is open to members of the public. Violent or peaceful forms of social protest that take place in open or closed public spaces such as streets, plazas, vicinity of prominent buildings fall in the category of demonstrations. Notable examples of demonstrations are as follows: Protest marches, silent marches, rallies, demonstrations, outdoor press declarations, gatherings, sit-ins, acts of civil disobedience, dharnas, bandhs, demonstrators, clashes with security forces, commemorative and religious processions (when they become means of protest, e.g. akhand paths), collective petitions (collecting signatures), self-immolation (and threats of self-immolation), hunger strikes, fasts, barricading, picketing (roadblocks), burning of vehicles, occupations of public buildings that are not included in the industrial action category, and the like.

#### *4.5.3.2 Industrial actions*

Industrial actions are types of protest events that take place within workplaces and/or involve the production process in the protest. Notable examples include any kind of strike (comprising slowdown strikes, wildcat strikes, sympathy strikes, and green bans), workplace occupations, boycotts, picket lines, and gheraoes. Note that actions that take place on public streets and roads, which are otherwise scenes of demonstrations, the previous category defined above, might become industrial actions when these spaces are actual workplaces of certain groups of workers who carry out the protest. An example of this would be public transportation workers carrying out a protest by blocking roads. While a roadblock would otherwise fall in the demonstration category, here it becomes a group of workers disrupting the work process by withdrawing their services and somehow "occupying" their workplace. On the contrary, the march of a group of workers on the streets to draw attention to their work-related demands must be considered a "demonstration" rather than industrial action. Similarly, occupations or disruptions of public institutions that are not worker protests, e.g. when university students occupy education facilities, are to be considered demonstrations, and not industrial actions.

### 4.5.3.3 Group clashes

Group clash events are instances of confrontation that stem from politicized conflicts (e.g. identity or economic interest-based, or ideological conflicts) between social groups such as fighting, lynching, ransacking, arson, any armed or unarmed clash between civilians of a political nature. Actions that target religious, ethnic, or similar minorities fall within this category. Said actions can be unidirectional, that is, the target of the event might or might not retaliate during the event. Group clashes that take place in open public spaces might involve or escalate from actions that are defined in the category of demonstration. The key to identifying group clashes would then be the actors involved. The distinguishing feature of group clashes is that both the participants (or organizers) and the targets of these events are non-state actors. While in violent demonstrations, the targets would be state institutions, state officials, and the like, group clashes target private individuals, businesses, religious buildings, etc. Also note that group clashes can be violent events that might involve the use of firearms, making these events somewhat hard to distinguish from armed militancy events described in the next category. Again, both participants and targets being non-state actors is a clue to identifying group clashes.

### 4.5.3.4 Armed militancy

Politically motivated violent actions that fall within our event definition are included in this category. These actions are usually carried out by political and/or militant organizations which systematically resort to violence as a means of reaching their goals. Their targets can be officials or civilians. Notable examples of this type of event are kidnapping, assassination, bombing, suicide bombing, hijacking, and the like. A feature of these events that distinguish them from group clashes is the nature of the perpetrators. Violent actions that target civilians will be included in this category if the perpetrators are armed militant organizations. Also, even though such actions are generally carried out by organizations, there might be cases in which such acts are carried out by individuals not necessarily affiliated with organizations or groups. Hence, if the act which intentionally resorts to violence carrying political goals, they are included in our definition even when they are carried out by a proverbial "lone wolf".

### 4.5.3.5 Electoral politics

Electoral politics events are rallies, marches, or any similar mass mobilization event that is organized within the scope of election campaigns of political parties or leaders. These mobilizations aim to garner support or ask for votes for the political entity competing in the upcoming elections. Note that demonstrations or protests that take place during the elections, such as boycotts, clashes, sabotaging voting, etc. will *not* be considered electoral politics and will be tagged with the corresponding semantic tag other than electoral politics.

### 4.5.3.6 Other

The "other" tag will be used for any event which does not fit in one of the categories listed above.

## 4.6 Event's Time and Location Arguments

### 4.6.1 Time of the event

Any information which gives information about the time the event takes place will be annotated with the event_time tag. The date information such as the year, month, or day; information about the time of the day as in "in the morning", at 3 p.m., "late at night"; less definite expressions such as "three weeks ago", "this year", "recent" are all valid event time expressions to be tagged. Keep in mind that the time of the event might be expressed more than once in different ways. All such expressions must be labeled as long as they denote the time of the event.

**Tag rules**

  i. Tag span rules:

  a) Time expressions will be tagged in their entirety, including the prepositions which indicate the time information as a whole. E.g. "at 3 pm", "before noon", "at the early hours of the day", "three weeks later", "in last September", etc.

  b) Complex time expressions that contain multiple values but indicate a single piece of time information will not be divided. e.g. "at 6 p.m. yesterday" "June last year". Similarly, expressions which indicate the day of the month will be included in the tag span. e.g. "Monday (15)"

  ii. Indirect and relative time values will *not* be tagged as event time. A case of this is the expressions regarding the frequency of recurrent events such as "75th anniversary" which are not be tagged as event time. Time expressions that are related to other events or developments such as "after the meeting", and "before the prime minister's visit" are also not to be considered as time information and will not be tagged as event time.

  iii. Duration expressions will not be tagged as event time information, e.g. "after 102 hours of hunger strike", "the strike ended in its fourteenth day".

  iv. The expressions which give information about the beginning and the end of the events which take place over long periods, such as strikes or hunger strikes, will be tagged as event time. e.g. "The strike ended on Monday"

### 4.6.2 Place of the event

Event places are official geographical or administrative divisions in which protest events take place. Each country has its distinct system of categorizing its administrative and geographical divisions. States, provinces, cities, districts, neighborhoods, villages, taluks, and/or panchayats can be counted as examples of these divisions. The given official place names of these divisions will be tagged with the event_place tag. A division might have local administration such as municipalities in cities or governors in states. Some divisions however will not correspond to an administrative unit but will simply indicate a geographical definition, such as districts, neighborhoods, or regions in some countries. The requirement for the label of event_place is official recognition and/or naming. Names of built environments such as streets, plazas, buildings, etc. are not to be considered place names; they take another tag, which is defined below.

**Tag rules**

i. Tag span rules:

   a) Only the Place name will be tagged. Generic settlement categories (village, city, etc.) and prepositions will *not* be included e.g. ==Chapadu== mandal, near ==Charar-e-Sharief==, in ==Bihar=='s impoverished ==Aurangabad== district, ==Gecek== köyü. Nouns such as town, village, etc. in these expressions will be tagged as location identifiers (See the section "Centrality").

   b) Direction expressions like east west will be included only when they are parts of the official names e.g. ==Imphal west== district, ==Salem West== district, ==West Bengal==.

   c) The conjunction "and" will only be included if it is part of the official name of the place, if it is used to connect different places, it will not be a part of the span and places it connects will have separate tags (this rule is valid for all the tags) e.g. ==Jammu and Kashmir==, ==Karnataka== and ==Gujarat== states.

   d) Prepositions that are preceding a place name should not be included in the event_place annotation. e.g. "Tens of thousands of workers went on strike in ==Hong Kong=="

   e) Expressions like "city" or "center" might be included in official place names; in which case they must be included in the event place tag span. The way to discern whether such expressions are part of official place names is by looking at whether they are written capitalized or not. We will assume that capitalized expressions like the "==Central== district of ==Hong Kong==" or "==Ciudad Autonoma de Buenos Aires==" (commonly abbreviated as ==CABA==, sometimes referred to as simply =="la Ciudad"==) are official event place names, and will tag them accordingly. In other instances where the expressions "city" and "center" are not capitalized, they will be excluded from tag spans.

ii. If the article mentions more than one place that the event takes place in, all place names must be tagged. This happens in two situations. The different scales of locality which have different names might be mentioned, as in "==Salem== city in ==Tamil Nadu== state". Also, events that start at one place and finish at another without interruption such as "Teachers marched through ==Kherwadi==, ==Bandra West== and ==Pali Hill== neighborhoods." Keep in mind that in other cases, events that take place at multiple, different places are to be treated as different events (see the next rule).

iii. If the article mentions separate events from multiple places, mark all events and the event-related places and note the relevant event number in the comments section (see the Event Separation Section).

iv. If a protest news report about a country within our scope contains protest events in another one that is outside of it, all the information about them will also be labeled. However, country names will never be tagged event places.

### 4.6.3 The facility in which the event takes place

Facilities are human-made structures or built environments that serve as the location the events take place in. These facilities can be streets, factories, airports, universities, vicinity of prominent buildings or sites (e.g. in front of a party office or an organization). Nouns that indicate the class of facility that the event takes place in will be tagged with the facility_type tag, e.g. "on the streets, in front of the court, towards the building", etc. The unique, named facility entities will be tagged with facility_name tag, as in "Safdarjung Hospital", "The Supreme Court", "INTUC headquarters", etc. Note that all facility expressions will be tagged with the corresponding prepositions and qualifiers which indicate the location information in full. An event taking place "at" a certain place has a different facility from another one that takes place "in" it. E.g. "the strike had crippled all services in the hospital", and "the strikers gathered in front of the hospital".

**Examples**

- Owners of contract buses, auto-rickshaws and taxis stayed off the roads.
- Shah's supporters also had gone on the rampage outside the principal's office on Tuesday.
- The strike had crippled all services in the hospital and most patients had been discharged.
- Two individuals died in the bomb explosions organized by radicals in places of worship
- Services in government hospitals were paralyzed due to the strike.

**Tag rules**

i. Tag span rule: The prepositions and other qualifiers which designate the location of the event in full, will be included in the tag span e.g. on the Osmania University campus, at Cusrow Wadia Institute of Technology, into the university campus.

ii. Like the event place, an event might take place in multiple facilities. As long as the event is continuous or the multiple facility references are of different scales, as in "in front of the bus stop, on the main street", all of them will be labeled separately with facility tags. In other cases, different facility references indicate different events which must be separated as per described in the Event Separation section.

iii. Take care in distinguishing between the targets and facilities of events. In cases where buildings or built environments are attacked, damaged, or targeted by the protesters in any way, they will be annotated with facility tags. Refer to the Target Characteristics section for a more detailed elaboration on this.

iv. In industrial action events such as strikes, the facilities will be the workplaces in which the strike is taking place.

v. Capitalization might indicate unique, named entities that are to be tagged with the facility_name tag, but do not rely solely on this. There are times when different news sources do not apply this principle consistently, also text extraction processes might change character cases. Instead, try to decide whether to use the name or type tag based on the facility that the article mentions being unique or not. If you cannot decide for certain, use the facility_type tag. Facilities such as municipality administration buildings, governors' offices, and the like are commonly unique.

### 4.6.4 Centrality of the event location (Urban or rural)

The linguistic information that relays information about the centrality of the place in which the event takes place will be labeled with urban and rural location identifier tags. The expressions such as "town", "city", "village", "panchayat", "hamlet", or "township" (South Africa), which directly indicate this, will be tagged. In addition, expressions that identify built environments (human-made surroundings as settings for human activity) such as plaza, avenue, slum, (village) well, that are invariably associated with urban or rural settings will be tagged with the corresponding location identifier tag.

**Tag rules**

i. The place name itself is not an urban or rural identifier. Nouns such as village, city, and town which follow place names will be tagged with urban or rural location identifier tags, as in Matahalli village, Vadadora city, Rocinha favela, Gecek Köyü, etc.

ii. Plaza, avenue, slum, and favela are typical expressions where urban location identifier tags can be used. Be careful to add more to this list and do not tag other entities with location identifier tags without consulting with the annotation supervisors. Types of buildings, institutions or types of economic activity, such as hospitals, farms, universities, etc. will not be tagged with location identifier tags.

iii. If expressions such as avenue or plaza are facilities of protest events and thus tagged as facilities, they will not be tagged with location identifier tags. In other words, there will not be overlapping location identifier tags and facility tags will have priority.

## 4.7 Event Actor Arguments

### 4.7.1 Participant Arguments

#### 4.7.1.1 Types of participants

The participants are the main actors of a protest event. They can be individuals or groups that engage in a protest. Groups may be referred to as masses (common people, public), armed or violent political actors (attackers, militants, militias, guerrillas), unarmed non-violent political actors (activists), blue-collar workers (factory workers, low skilled workers) white-collar workers/professionals (lawyers, doctors, nurses, corporate employees, teachers, engineers), the unemployed, students, women, LGBTI Individuals, peasants (including landless peasants and sharecroppers), villagers, war veterans and their relatives, politicians (members of the parliament, political leaders, party members), academicians, journalists, public figures, artists, intellectuals, artisans and shopkeepers, the disabled (physically challenged), prisoners and their relatives, refugees/immigrants, consumers, hackers or hacktivists, ultras, sports spectators, supporters, relatives of political violence victims, prisoners, kidnapped, disappeared or assassinated individuals, and the like.

**Examples**

Teaching and non-teaching staff went on strike on Tuesday.

Hundreds of karamcharis staged a dharna in Delhi.

Skirmishes occurred between the Bharatiya Kisan Union (BKU)-EKTA activists and the police at various places in the district, on Monday.

Senior Congress leaders including Kamal Nath, Oscar Fernandes, Narendra Nath, DPCC chief Subhash Chopra and party legislators participated in the demonstration.

More than 800 agitators participated in the protest.

**Tag rules**

i. Tag span rules

a) We aim to include the fewest possible words that indicate the actual participants meaningfully. No qualifiers will be included in the tag span e.g. rival party supporters, resident doctors, Muslim women, jobless diamond workers, secondary school teachers, inter-city bus drivers, angry karamcharis.

b) As an exception to the previous rule, some qualifiers which modify certain occupational groups change the nature of those occupations so significantly that they become a different category of employees with different socioeconomic status and/or organizational characteristics. In these instances, the qualifiers should be included in the tag. It is not straightforward to determine these occupations as they require certain unique country characteristics to be taken into account. When a likely case is encountered, don't hesitate to ask the project experts about the coverage of the tag spans. Prominent examples are as follows: auto-rickshaw drivers, college teachers, safai karamcharis. Note that the words, drivers, teachers, and karamcharis are the only words to be included in the tag spans in other instances, see the previous rule.

ii. Words like mob, group, and crowd, will be tagged only if they are the only words standing for participants. When they are parts of phrases with more definite participant words, they will not be included e.g. a group of about 100 people, group of locals, group of militants, group of students, the angry mob, the crowds gathered in front of the supreme court.

iii. Named entities or expressions referring to a specific group of people may be expressed with different expressions after their first occurrence. The different occurrences should be annotated with the tag(s) used in their first occurrence if the alternative expression also denotes the same class of entity (event type, participant type, sector, etc.). E.g. If a tagged expression such as "teachers' strike", where the word "teachers" is tagged with participant_type and participant_SES, is followed by a phrase such as "agitators want a higher salary" where "agitators" refers to teachers, the word "agitators" is to be tagged as participant_type only as the word "agitator" denotes a type of participant but denotes neither the sector nor the SES.

iv. Groups of people regardless of whether they are organizers, participants, or targets may be expressed with pejorative terms, e.g., "miscreants". This type of a term should be annotated with the tag corresponding to its role in the event, that is, organizer, participant, or target.

### 4.7.1.2 Names of participants

The name of the participant(s) may be mentioned. This can be the first name and/or the last name of an individual who is engaging in the protest. Please use the tag participant_name for annotating these names. Titles such as Mr., Ms., or Dr. will not be included in the tag span.

**Example**
- Senior congress leaders including ==Kamal Nath==, ==Oscar Fernandes==, ==Narendra Nath==, DPCC chief ==Subhash Chopra== and party legislators participated in the demonstration.

### 4.7.1.3 Participant Count

Any expression that indicates the total quantity of individuals who participated in the event should be annotated with the tag participant_count.

**Examples**
- ==a large number== of people, many people
- ==hundreds== of BJP workers, four agitators
- as many as ==50== employees gathered at the DC's office
- more than ==800== agitators participated in the protest
- about ==700== members of the BKU are sitting on a dharna.

**Tag rules**
i. Tag span rules:
   a) In Expressions, such as "more than a ==hundred== demonstrators", "nearly ==fifty== assailants", "as many as ==five thousand== workers", and "about ==twenty== militants", that give an estimation of the number of participants, only the number expressions will be tagged; that is the estimation qualifiers will not be included in the tag span.

b) Complex participant count expressions which contain multiple values, but indicate a single participant quantity information will not be divided. e.g. "between 800 and 1000 people marched in the streets"

ii. Sometimes quantity expressions refer to a part of all participants (e.g. the police arrested 11 of the protesters). These expressions should not be tagged with participant_count as they do not give information on the total quantity of the participants.

### 4.7.1.4 Ideology of the Participants

In case an explicit mention of participant ideology is mentioned, this is tagged as participant_ideology. The participant ideology can be communist, socialist, anarchist, social democrat, left-wing, right-wing, feminist, liberal, nationalist, religious fundamentalist, conservative, far-right, green or environmentalist, animal rights, pro-LGBTI, and any expression similar which identifies the ideology or world view of participants. Please tag any word/phrase that indicates the ideology of the participants.

**Examples**

- Left-wing militants blocked the road in Caracas.
- Maoists staged a rally in the city center.
- The city was the scene of massive anti-government demonstrations.

### 4.7.1.5 Ethnic identity of participants

Expressions that identify the ethnic and racial identity of participants are marked with participant_ethnicity.

**Examples**

- Catalan protesters took to the streets of Barcelona.
- Two of the activists who participated in the event were identified as Tamil militants

### 4.7.1.6 Religious Identity of Participants

Any expression that identifies the religious affiliation of participants is tagged with participant_religion.

**Examples**

- Some Catholics are protesting what they regard as a blasphemous stage production that disrespects God and the Catholic faith.
- Nationalistically aroused Sikhs are leading demonstrations to ban the sale of cigarettes since the use of tobacco is forbidden to Sikhs.
- Muslim workers protested against the ban of Friday praying in factories in Mumbai.

### *4.7.1.7 Caste of the Participants*

Expressions that designate the caste of participants will be annotated with the participant_caste tag. This tag is specific for the annotation of news from India.

**Example**

They also explained the aggression of the nayaks on harijans on every trivial issue.

### *4.7.1.8 Socioeconomic Status of the Participants*

The expressions which give information as to the social standing or class of participants will be tagged with participant_SES. Any word or phrase that indicates the class position, socioeconomic status, income, or education level of participants should be annotated.

**Tag rules**

i. The expressions which designate, if indirectly, the socioeconomic status of participants, will be tagged. Examples can be "starvation", "malnutrition", "slums", gated community, etc.

ii. If the participant type is a category implying participant SES (as is the case for teachers, doctors, workers, etc.), mark them with both participant type and participant SES tags. The words "politicians", "activists", "students", and "housewives" do not imply any SES information and hence will not be tagged participant_SES.

### 4.7.2 Semantic Categories of Participants

The research agenda of the EMW project involves examining social groups and classes which engage in contentious politics actions, because how governments respond to protest events is determined by who engages in protest action in important ways. Thus, rhe sociological characteristics of the participants of contentious politics events are as important as the characteristics of those events themselves. Just like protest events, the GLOCON database aims to classify protest participants into sub-types. The semantic categorization of participant actor expressions serves this task. Since participant names give little linguistic information that can be classified, semantic category tags will be put on expressions that are labeled with the participant_type tag.

**Tag rules**

i. Every expression that is tagged with participant_type, must also have a participant semantic category tag.

ii. Semantic label tag spans will be coterminous with those of the participant_type tags that they are used in an overlapping manner.

iii. Participant semantic tags will be consistent for each expression that refers to the same group in different event sentences throughout the text. In case, for instance, a group of "teachers" who go on strike is referred to as "strikers", "protesters", or "agitators" in different sentences, every such expression will be labeled with the same semantic tag ("workers" in this example).

The tags for participant semantic categories are as follows:

### 4.7.2.2 Peasant

People who work in agriculture and/or live in rural areas. The category includes smallholders, that is subsistence farmers who own small plots and use mainly domestic labor.

### 4.7.2.3 Worker

Any kind of blue or white-collar employee in private sector companies or public institutions. This category should exclude those who are classified under "professionals".

### 4.7.2.4 Small producer

Owners of small shops, small traders, artisans, or any small business which employs family labor (including transport owners like 4 owner taxi drivers).

### 4.7.2.5 Employer/executive

Owners and managers of medium and large size businesses that employ waged labor.

### 4.7.2.6 Professional

University-educated professionals who are self-employed or work in the private or the public sector (physicians, lawyers, academics, journalists, etc.).

### 4.7.2.7 Student

Students from all levels of education.

### 4.7.2.8 Politician

Members of political parties who hold office in legislative organs and/or executive branches of government. (There may be rare cases in which politicians are independent, i.e. not affiliated with political parties. Party membership, although very common, is not mandatory.)

### 4.7.2.9 Activist

Ordinary members of political parties, and members of grassroots organizations and NGOs. In distinguishing activists from ordinary people who engage in protest, the key here is to look at whether in the news report, participants are organized under a political organization. In most cases, activists will be mentioned together with the NGO or party that they belong to. In the absence of such an explicit mention of an organization, expressions that denote activism will be used (e.g. environmentalist, activist, feminist, human rights advocate, etc.). Note also that political party members who hold offices in governments and parliaments should be labeled as "politician" rather than "activist".

### *4.7.2.10   Militant*

This category contains members of political organizations which are commonly illegal and resort to armed violence (e.g. Islamic fundamentalist militants, members of armed revolutionary organizations, etc.). In some cases distinguishing militants and activists will be difficult. There will be events where certain members of ordinary, legal political parties will resort to armed violence. On the other hand, in certain events, you will see members of militant organizations engage in peaceful forms of protest. The way to decide in such situations is to consider the character of the political organization with which participants of these events are affiliated. Thus, we will decide based on how the organization is characterized in the press, rather than the type of the event.

### *4.7.2.11   People*

This is a broad category that refers to citizens without organizational or employment ties. Examples include women, residents, religious or ethnic community members, etc. Non-specific expressions which are labeled as the participant_type such as mob, crowds, etc. will also be labeled as people.

### *4.7.2.12   Other*

Any type of participant that cannot be placed into the categories above. This can be because the given participant expression is not covered in any of the categories above, or it is unclear to which type of protester is the expression referring.

## 4.7.3   Organizer Arguments

Organizer is another actor category of actors that engage in contentious politics. Organizers organize and/or lead the event and participants. Some events do not have organizers because they are spontaneous events, in which participants take action without being led by anyone. In other cases, an organization, institution, a group of people, or a single individual might organize or lead the protest. Any organization which takes part in the protest and individuals who lead or organize the event will be tagged with organizer tags. Individuals can be the organizer only in cases in which it is explicitly stated that they lead or organize the event.

### *4.7.3.1 Distinguishing between organizers and participants*

The difference between the organizer and the participant of an event has important implications. Protest events of a particular class or participants will likely have very distinct characteristics depending on the organizers. Also, that some events will not have explicit organizers, i.e. will likely be spontaneous agitations is an important detail that we need to know. However, as both are authors, or main actors of events, deciding who is participating and who is organizing might prove tricky. Here, the distinction between these two classes of actors will be defined in detail. To begin, we can hypothesize that any event will have one of the following organizer-based characteristics:

➔ There is not any explicit organizer, spontaneous events

- There are organizers and participants, i.e. the event is coordinated. (Unions leading a strike, for example.)
- There are organizers, but there are not any participants: This is not very common. We see some parties/organizations coming together to organize an event, in which no one participated.
- There are no organizers and participants mentioned. This could be an explosion, for example, where no one was identified as responsible.

Some examples that clarifies this distinction are as follows:

- In case there is a boycott at a university campus, the organizer is the student union and the participants are the students.
- The potential organizer of a strike is a labor union and the participants are the employees.
- Some events are reported as led by a particular person, in such a case the person is the organizer, e.g. "the protesters led by Shelia Dikshit", where "Sheila Dikshit" is the organizer_name. If the person entity is not explicitly specified as leading the event, it should be annotated with the appropriate participant tags. For instance, an assassin is still a participant in an assassination event.
- A phrase like "association members" should be tagged as "association": organizer_type and "members": participant_type.
- Some cases mention the names of both an organization and its leader, as in "BJP leader Madan Lal Khurana, architect of many anti-government protests, lent his voice, and a broad smile." Here, "BJP" should be labeled organizer_name, and "Madan Lal Khrana," participant_name.
- In cases where organizations or individuals express their support for a protest event without actually participating in it, they should not be labeled with participant or organizer tags.

### *4.7.3.2 The event organizer type*

The organizer(s) of an event can be Political parties, non-governmental organizations, associations, charities, trade unions, non-party political organizations (students' organizations, women's organizations, etc.), Anti-state armed groups, such as terrorist organizations, resistance movements, guerrilla groups, pro-state armed groups such as paramilitary organizations, Social communities (tribes, religious groups), Local public authority (municipalities, governorates, etc.) or Multiple organizations (more than one party, NGO and another type of organization coordinated the event). Please tag any type of the organizer with organizer_type tag.

**Examples**

- Left parties organized a silent march.

- A prominent banned religious outfit, Deendar Anjuman, exploded a mine on Sunday

**Tag Rules**

i. Tag span rule: Like any other tag span, the shortest possible type expression will be tagged, i.e. words like association, party, union, employees association. If the word association stands for a workers union, which is generally the case in India, the word before the association will be included, i.e. employees association.

ii. Business entities, companies, and firms are civilian actors which can engage in conflict with other civilian actors, be it their workers or residents of the localities where their plants are located. Most often, companies are actors that are at the receiving end of protest events, i.e. they are the targets of protest events engaged in by their adversaries mentioned above. However, in certain conflictual situations, their security or other personnel might engage in clashes or similar confrontations with other civilian actors. When the news report refers to "company" "firm" or similar entities as acting subjects of protest events, they will be labeled as organizers. Expressions that refer to individuals that are associated with companies, such as security personnel, will be labeled with participant tags. e.g. In the sentence "The firm retaliated to the protesters by burning their tents", "firm" will be labeled as organizer_type tag.

### *4.7.3.3 The Name of the Event Organizer*

In case the name of the organizer of the event is mentioned, please tag it/them with the organizer_name tag.

**Examples**

- "Frankie Boyle has started a hunger strike in solidarity with Guantánamo Bay prisoner Shaker Aamer",
- "The Labour Party organized a rally on Sunday",
- "A prominent banned religious outfit, Deendar Anjuman, exploded a mine on Sunday"
- The University Teachers' Association joined the campaign and signed the anti- austerity statement.
- "...when the protesters, led by senior youth congress leader Ish Kohli, held a demonstration just before the beginning of the council meeting, Firdous came down and received a memorandum demanding not to handover the property to the private organisation". youth congress: organizer_name, leader: organizer_type, protesters: participant_type, Ish Kohli: organizer_name.

**Tag Rules**

i. Tag span rule: Full organizer name will be included. If the abbreviation of the name is present within brackets, it will be tagged separately without the brackets e.g. Datta Gule's auto union, All India Bank Officers' Association, (PUCL), Indian National Trade Union Congress (INTUC), Joint Front of Trade Unions of the GIC and the LIC.

ii. As it was mentioned in the Facility Name section, capitalization might indicate a unique entity that must be tagged with an organizer_name tag but due to possible errors in syntax or tokenization, this should not be relied upon on its own.

### 4.7.3.4 The Ideology of the Event Organizer

The organizer of a protest may have an explicit ideology, which can be at least one of communist/socialist, anarchist, social democrat, mixed left (including feminists, environmentalists, etc.), other left-wing ideologies, conservative, liberal, nationalist, religious/conservative, far-right, other right-wing ideologies, mixed right, feminist, green movement/environmentalist, animal rights, pro-LGBTI, mixed right/left, etc. Please annotate the occurrence of an ideology with organizer_ideology.

**Example**

- Civil rights organizations and Left parties on Friday staged a protest rally

### 4.7.3.5 The Ethnicity of the Event Organizer

If there is reference to the organizers' ethnic or racial identity, label the identity with organizer_ethnicity tag.

### 4.7.3.6 The Religion of the Event Organizer

The organizer's religious identity is annotated with participant_religion.

### 4.7.3.7 Caste of the Event Organizer

In case the caste of the organizer is provided, it should be annotated with the organizer_caste tag. This tag is specific for the annotation of news from India.

### 4.7.3.8 Socioeconomic Status of the Organizer

If the news article contains information about whether the organizers of the event are poor, rich, or middle-class people annotate this with the tag organizer_SES. Please annotate all mentions.

**Tag rules**

i. If there are SES indicative phrases are related to organizers, mark them with this tag. These can be "starvation", "malnutrition".

### 4.7.4 Semantic Categories of Organizers

We classify organizations that engage in protest into different sub-types, as we do the participants. Hence, every expression which is labeled with an organizer_type or organizer_name tag will be labeled with an organizer semantic category tag as well. Note that different from participant names, organizer named entities still contain generic organizer type information, which is the reason why we label organizer names as well as organizer types.

**Tag Rules**

i. Every expression labeled with organizer_type and organizer_name tags will also be labeled with the organizer semantic tags (see overlapping annotation section).

ii. Organizer semantic tag spans will be coterminous with those of the participant_type tags that they are used in an overlapping manner.

iii. Organizer semantic tags will be consistent for each expression that refers to the same organizer in different event sentences throughout the text.

The tags for organizer semantic categories are as follows:

### 4.7.4.2 Political party

As the quintessential and ubiquitous political organization, political parties are the most common organizers of contentious politics events. Every organization which refers to itself as a political party to itself will be labeled with this organizer semantic tag regardless of it competing elections, take part in government, and even being legally recognized as such. The only exception to this are armed/militant organizations that refer to themselves as political parties, which will be labeled with militant/armed organization tag (see below).

### 4.7.4.3 NGO

Any non-governmental, grassroots, activist, or charity organization will be labeled with the NGO semantic tag.

### 4.7.4.4 Union

Trade unions, i.e. organizations, associations of blue or white-collar workers which aim to protect their rights, and improve the conditions under which they work. Unions are usually organized based on economic sector or activity. They are usually legally recognized and protected organizations, but there will be instances when such recognition is not present and workers would be acting together under the banner of a union even if it is informal. This category is closely associated with the "industrial action" event category, i.e. in most cases unions will be organizers of protest actions that are considered under that heading. Therefore, mentions of "social movement unions" should be considered under the NGO category, rather than union.

### 4.7.4.5 Militant/armed organization

Political organizations that are illegal and routinely resort to armed violence in achieving their goals will be labeled with this tag. In most cases, the use of armed violence is built into the founding principles of these organizations, which commonly identify themselves within an armed conflict with states and/or other political organizations. In other words, when otherwise non-violent political organizations (e.g. ordinary political parties) engage in occasional violent (even armed) protest actions, this is not a sufficient basis on which to classify them as militant organizations. Similarly, reference to non-violent political organizations as "terrorists", etc. by governments or news reports is not enough on its own to classify them as militant organizations.

### 4.7.4.6 Chamber of professional

This is a special category of economic organization which brings together practitioners of certain legally protected professions, such as medical, or legal professions. This is different from unions in that chambers have legally recognized regulatory functions in the conduct of their professions, such as licensing practitioners, determining fees, etc. There are variations in different countries as to how these organizations are established and operate. An example of this type of organization, which is found in many countries, is bars, i.e. institutions of the legal profession. In some cases, medical associations and associations of engineers have a similar form. While these organizations are not usually politicized, in some countries, they can become prominent organizers of protest events for their members.

### 4.7.4.7 Person

When individuals are labeled as organizers of protest events (see the organizer_type section for the cases that this is so), they will be labeled with the "person" organizer semantic tag. Note that this tag will be used with both the organizer_type expressions (such as "leader", "president", etc.) and named person organizers (such as "Lula", "Kirchner", etc.).

### 4.7.4.8 Other

Any expression that is labeled with organizer_type or name tag which cannot be placed into categories defined above will be labeled with the "other" organizer semantic tag.

## 4.7.5 Target Arguments

The organization(s), institution(s), group(s), or individual(s) that is the antagonist of the protest event is annotated with the target tags. Targets are the entities that are opposed by those who organize or participate in the protest. Certain protest actions can aim at the targets directly for instance when they are attacked by the protesters. In other instances, the entities towards which the actions are directed are not the targets but the objects of the protests. In the latter case, care should be taken not to tag the objects of the protest as targets. Consider the examples where "small businesses closing their shops to protest some ==government== regulation," "workers blocking roads to protest their ==employer==." "a majority caste members attack the settlements where members of a ==minority caste live==. The entities italicized are the objects of protest actions but the true targets are those expressions that are highlighted. In brief, the target implies the entity that is opposed by the protesters in the abstract. Most of the time, the objects of the protest events are entities annotated with facility tags. Facilities, as defined above, are human-made structures or spaces in which the protests take place, and should not be confused with targets. In the examples already mentioned, shops, roads, and settlements are facilities even though the actions are directed at them; whereas the government, the employer, and the minority caste are targets, i.e. the antagonists of the protest.

### 4.7.5.1 Target Type

The type of entities that are the targets of the protest events, such as organizations, individuals, groups countries, etc. should be identified and annotated with target_type. Examples of the organizations are Government (governing party and its politicians, ministries), state, the army, police, employer & business, mayor or governor (including local governments, municipalities, and the governorate), Other political parties and politicians (except for the governing party), judiciary, civilians and non-political social groups (students, women, workers, etc.), minority groups (ethnic, religious, racial, caste, etc.), non-party political organizations (including non-state armed groups, activist organizations), political activists, politically affiliated individuals, intellectuals and alike, public institutions, bureaucracy and bureaucrats (regulatory agencies, public schools, and hospitals), the parliament, foreign countries, NGOs, transnational organizations, etc., unions, armed paramilitary organizations, criminal organizations (drug trafficking, mafia, etc.). In case the event did not occur against a specific individual or institution or it is not clear what the target is, there will not be any annotation.

**Examples**

- During the protest, activists accused the minister of violating the international law.
- More than 2000 persons staged a protest near Vijay Theatre ground against the district police.
- Contract Bus Association chief Harish Sabharwal was booed by the protesters.
- Priests were among the 7 that were shot dead in Kashmir.

### 4.7.5.2 Target Name

The name of the target should be annotated with the target_name tag.

**Examples**

- NATO, World Bank, Prime Minister Theresa May
- Contract Bus Association chief Harish Sabharwal was booed by the protesters
- After 102 hours of hunger strike, there has been no effort from the ADCB chairman to resolve the issue.

## 4.8 Event Separation

### 4.8.1 Introduction

Sometimes news articles include information about multiple events at once. Pieces of information about one event might not apply to the other event. In this case, we need to distinguish the events within the article from each other. This is referred to as event separation and is subject to several rules to ensure coherence in coding. Note that in separating events we need to think of the news text rather than the actual reality that the text recounts. That is to say, we are more interested in the separate event references in the text than whether the said events are separate from each other in real life. As will be clearer in the examples below, sometimes it is not possible to know or show for certain whether distinct event references correspond to distinct real-life events.

We use comments in the tag window of the FLAT environment to mark the distinct event references in the text. Each event will be numbered from 1 and on, following the order in which they appear in the article. The comment style will be as in "Event 2", "Event 2, Event 3" for two events, or "Event 3, Event 4, Event 6" for multiple events, etc. The first event will not have an event number. If the tags do not contain an event number comment, they will be assumed to be related to event 1.

Every tag (meaning all tags from all annotation foci) that belongs to a separate event will contain the same event number in the comment.

**Example**
- "At noon, BJP workers gathered in the square and shouted slogans, condemning the failure of the Union Government in delivering justice to the victims of last year's terror attack at the train station where armed militants killed 25 people."

There are two distinct events in this sentence. BJP workers' demonstration is the first event, and the attack at the train station is the second event (in the order in which they appear in the document). The tags of the first event will not include an event number comment and are as follows:

At noon: e_time; BJP: org_name; workers: part_type; gathered: e_type, demonstration; in the square: f_type; shouted slogans: e_mention ,demonstration; Union Government: target_type.

The second event's tags will all contain the comment "Event 2" and are as follows:

last year's: e_time; attack: e_type, armed militancy; at the train station: f_type; militants: part_type; killed: e_mention, armed militancy.

### 4.8.2 Situations in which events will be separated

The separation of event references will be based on the difference in at least one of the following: event time, event place, facility (name or type), or participants.

**Time difference**

Events that occur at different times will be separated from each other. The time difference necessary for separation is 24 hours. Events that continue throughout the same day will not be separated even if they are reported to occur at different times of the day.

### Location difference

Events that are reported to take place in different locations will be separated as distinct events. Locations can be event places or facilities. An event that has started at one location and continued at another, e.g. a march that started at one location and proceeded somewhere else, will be counted as one event despite multiple location references. However, if an event is happening at multiple locations (simultaneously, or discontinuously), it will be divided such that every location reference will count as a separate event. Likewise, multiple event references (i.e. plural event mentions such as protests, attacks, etc.) will be divided such that both the plural mention and following singular mentions will be counted as distinct event references (see below Plural Event Mentions for clarification).

### Participant or organizer difference

Events that are carried out by different participants or organizers with separate goals and motivations will be separated. This separation will take place even in cases where different protests occur at the same time and location. The separation is based on event motivations or goals but since motivation info is not something that we annotate and might at times be elusive, we try to distinguish events based on participants and organizers. The case most frequently exemplifies this situation is counter-protests, where two groups of participants or organizers demonstrate against each other and/or with conflicting agendas. Note that in cases where there are multiple types of participants and/or organizers that protest together (i.e. participate in the same protest), the event will not be divided.

### Semantic event category difference

Events that occur at the same time, same location, are organized and participated by the same actors, but have a different semantic event category will be divided into different events for each distinct semantic category. Put differently, the references to each distinct event in a document have to have the semantic category tag. Although this case is rare, it can be encountered when rallies, marches, or other types of demonstrations occur during industrial strikes.

## 4.8.3 Rules of event separation

i. The rule against annotation outside event sentences applies in the case of separated events. In a multiple event situation, expressions that give information about an event whose event reference (event_type or event_mention) is not in the same sentence, will not be labeled, even if the sentence contains a reference to another event, i.e. is an event sentence of another event. Consider a slightly different version of the previous example:

- "At noon, BJP workers gathered at the train station and shouted slogans, condemning the failure of the Union Government in delivering justice to the residents of Mumbai.

  The incident which had taken place last year at the train station where armed militants killed 25 people and the manner in which the government handled the situation has been a chief agenda item for BJP ever since."

In this example, the "at the train station" expression is the facility argument of both the BJP gathering and the militant attack. Yet, in each sentence, it will only contain the event number of the event reference that is copresent in the sentence. Also, note that the expressions "government" and "BJP" in the second sentence are not tagged due to the no tagging outside event sentence rule as they are not related to the "killed" event in that sentence.

ii. Plural Event References: In cases where the articles contain multiple events, there might be event triggers (types or mentions) that are plural, i.e. refer to more than one event that should be separated. We have a unique procedure for separating these events. In a nutshell, if an article contains a plural event reference, such as "protests" which refers to 2 different events, each of which is reported in the article, that article will have 3 separate event numbers.

Consider the example of the first three sentences of a document below:

- Karnataka State Government Employees Association engaged in a wave of demonstrations across Karnataka yesterday, urging the government not to go ahead with the new retirement scheme.

  In Bangalore, hundreds of workers participated in the rally in front of the collectorate.

  Mysore was also the scene of protest as around 3000 employees took to the streets.

The three sentences above will be treated as containing three separate event references, even though in real life the first reference "demonstrations" comprises the events in the following sentences. The reason for this is that the protests that take place in Karnataka, Bangalore, and Mysore all have different event place information attached to them and it is not possible for a reader uninformed beyond what is present in the actual news article, to decide for certain whether these three events coincide or not based solely on the place names. This also applies to cases where the plural events happen at distinct times and facilities (i.e. the case where demonstrations in the above example happen at two distinct times "within the last month", say, yesterday and 2 weeks ago. Again, there will be 3 event annotations: the first is the plural that happened "within the last month", the second is the one that happened yesterday, and the third is the one that happened "2 weeks ago".)

If the different event times and/or locations are mentioned in the same sentence, both the plural event trigger and other pieces of information belonging to the separate events will be numbered in the same sentence.

- "Karnataka State Government Employees Association organized demonstrations in Bangalore and Mysore yesterday, urging the government not to go ahead with the new retirement scheme."

Here, all tagged expressions except event places will contain the comment "Event 1, Event 2," as all of them give the same information about two events that are separated according to the separation rule based on distinct event-place. Bangalore will not contain an event number tag (because it is event 1) and Mysore will contain the tag "Event 2".

# 5  Violent Protest Events Annotation Manual

This task will follow the token level annotation task, although it is a classification task that is performed based on the whole document. The articles which contain at least one violent protest event, as per defined below, will be labeled as **violent**. On the contrary, the documents which do not contain any violent events will be labeled **non-violent**. The documents selected for labeling in this task are those that have already been identified as containing at least one contentious politics event, as defined in DOLPAM. Only those events which are included in our event definition will be evaluated based on violence. That is to say, events that do not conform with the event definition laid out in DOLPAM will not be evaluated and labeled.

## 5.1   Identifying violent protest events

Violent protest events are contentious politics events whose participants deliberately resort to violence at any time during the event. Violence refers to causing harm and/or damage to individuals or property in any way. The harm or damage in question must be inflicted deliberately, that is, accidents or any harm caused unintentionally will not be enough on their own to label the event as a violent one. Violence must be understood as any action which aims at:

   a) Infliction of bodily harm to any individual(s). Possible targets of the violent act include state officials (including security personnel), any civilian or group of civilians that are the antagonists of the protest (i.e. targets of the protest action), participants of the event themselves (in cases of internal clashes among participants, and acts of self-harm such as self-immolation, hunger strike and the like), or individuals which have neither any role nor relevance to the event such as bystanders, etc.

   b) Infliction of damage to property and/or environment: Government offices, public built environments (e.g. roads, parks), buildings belonging to any non-official entity (party or union offices, shops or homes belonging to a particular minority), vehicles, religious sites or places of worship, natural environment (e.g. in cases of forest burning), etc.

### 5.1.1  Some protest events that are violent

   i. Forceful removal or destruction of movable property such as ransacking, and looting are to be considered violent events. Theft and burglary actions serving political goals and/or carried out by political organizations, which are otherwise petty crimes, are also to be marked as violent. Examples of this are bank robberies, looting warehouses, hijacking public transport, extortion, mugging, and the like.

   ii. Nude parade is a lynch-like form of violent action where the victim is stripped naked and paraded in public spaces for degradation and insult. It is thus a violent event.

iii. Violence threats and attempts that are included in our event definition as per DOLPAM are to be considered violent events even if they are not carried out. Instances of hijacking commonly involve a threat of violence and thus will be considered violent events.

iv. Protests which involve deliberate self-harm by participants are considered violent events. Self-immolation and hunger strikes are among examples of self-harming protests.

v. References to events that are by their nature likely to involve violence must be considered carefully. Riots, civil unrest events and the like might be reported without explicit mention of the damages caused by violence. An event mention such as civil unrest or disobedience might not, on its own, be enough to label the event as violent. In such a case other details in the article, for instance, expressions such as "victims", "violence", and "damage" can give away the occurrence of violence. Riots on the other hand can be assumed violent on their own unless the article explicitly states that it is non-violent.

### 5.1.2  Some protest events that are not violent

i. Violence must be resorted to by the participants of the protest. In cases where there is violence directed at the participants by security forces but there's not an explicit mention of participants resorting to violence in retaliation, etc., these documents must be labeled non-violent. On the other hand, a protest event that was intended and started as non-violent but later involved participants' violent retaliation to state forces' attempt at its suppression should be labeled a violent event.

ii. Sometimes the news articles might use the expression "violent" while describing protest events without giving specific details as to the nature of the violence (e.g. "violent reaction", "violent opposition", "violently objected", etc.).  In such cases, the qualifier violent might refer to the intensity of the protest and thus not indicate actual violence. Take care in such cases to detect other expressions which leave no doubt as to the use of physical violence by the participants. If such statements do not exist, do not label the event as violent.

iii. Actions that can be characterized as symbolic violence which aim at mockery, insult, ridicule, defamation, or desecration are not to be labeled violent events. Such actions might be directed at sacred religious symbols, and artifacts of sentimental value such as flags, or persons (e.g. politicians). Examples include flag-burning, offensive slogans, paint bombs, throwing eggs, tomatoes, shoes, etc.

# 6 Protest Event Demands Annotation Manual

In the Emerging Markets Welfare project, we are interested in determining the effect of contentious political movements and events in shaping the welfare policies of governments that respond to them. Protest events have various demands and grievances that their participants enunciate at their core. In researching the effects of protest events and movements on welfare policies, a key question to address is whether the types of demands and grievances that give rise to protest events make a difference in terms of governments responding with welfare policies. In this task, we will classify news articles about protest events by the demand and/or grievance which motivates the protest action. To underline, similar to the violent events annotation task, the demand annotation task is also a classification task performed on the document level. For this classification, we define three broad categories of demand/grievance types in terms of their economic and welfare-related nature. These categories are as follows:

## 6.1 Non-Economic Demands

These are demands and grievances which stem from social and political issues that are not related to social conflicts based on the distribution of resources or production processes. Demands and grievances that stem from social and political conflicts based on identity and territory are to be considered within this category. Protest events that are motivated by issues regarding political representation, participation, human rights and liberties, and the environment are also within this category. To enumerate, all protest event demands and grievances that are based on ethnic, racial, gender, religious, and caste divisions; ecological concerns; political issues of territorial and/or sovereignty issues, representation, accountability of officials, corruption, state and/or social violence, political rights and liberties (such as freedom of expression), are to be regarded in the non-economic category.

## 6.2 Economic Non-Welfare Demands

These are demands and grievances that relate to social conflicts that are based on distribution and production relations with the exclusion of those that relate to social welfare, which is identified below. All protest events that are motivated by demands and grievances based on labor issues and other economic interest based conflicts, poverty and economic inequality, public infrastructural investment, and land ownership are to be categorized within economic-non-welfare category. To enumerate, issues such as cost of living, inflation, wage levels, labor protection and employment, taxation, infrastructure and municipal services, housing, and land ownership are to be regarded in the economic-non-welfare category.

## 6.3 Economic Welfare Demands

These are demands and grievances that stem from the issue of economic distribution and relate directly to social welfare. By social welfare, we refer to politics of social policy in its following dimensions: social security (old-age pension, unemployment and sickness insurance, non-wage benefits of waged employment), social assistance, education, healthcare, social care services (child care, care for the sick and elderly). Protest events whose participants demand welfare provision and/or higher quality welfare provision in any of the fields enumerated above are to be regarded in the economic-welfare category.

## 6.4 Rules of Demand Annotation

i. There will be news articles that will not report event demands explicitly. In these cases try to infer the demands from implicit information, for instance from the event types or background information present in the text. However, try not to base your decisions on information external to the news text.

ii. Strikes (industrial actions) will be labeled economic-non-welfare unless a welfare related demand is mentioned in the article.

iii. Demand annotation must be exclusive. If a protest event enunciates multiple demands, chose the first demand that the report mentions as the demand of the protest event.

iv. Similar to the rule above, documents that contain multiple separate events will be evaluated in terms of the demands of the first appearing event.

# 7 Main Publications Based on the Annotation Manual

# 8 Version History

## 8.1 Version 1.01 (20220511)

- A title page was added.

- Acknowledgments section was added.

- List of publications was extended.

- Sections 4.7.3.1, 4.8.2, and 4.8.3 were rephrased for clarification.

- Lots of minor edits were made throughout the document correcting language errors, typos, and punctuation.